\documentclass[11pt]{article}
\usepackage{placeins}

\newcommand{\blind}{0}

\usepackage[margin=1in]{geometry}
\usepackage{graphicx}
\usepackage{xcolor}
\usepackage{url}
\usepackage{pdflscape}
\usepackage[authoryear]{natbib}
\usepackage{tabularx}
\usepackage{longtable}
\usepackage{booktabs}
\usepackage{array}
\usepackage{makecell}
\usepackage{float}

\usepackage{subcaption}

\usepackage{algorithm}
\usepackage{algpseudocode}
\usepackage{amsmath,amssymb}
\usepackage{enumerate}

\begin{document}
		
		\def\spacingset#1{\renewcommand{\baselinestretch}%
			{#1}\small\normalsize} \spacingset{1}
		
		\if0\blind	
\title{A Data-Driven Framework for Unsupervised Monitoring of Transmission Systems Using End-of-Line Testing Data: A Case Study at Ford Motor Company}
\author{
Mohammad N. Bisheh$^a$, Mehrdad Moradi$^a$, Parinaz Farajiparvar$^b$,\\
Colin Brady$^b$, Rajesh Gupta$^b$, Xueling Li$^b$, Javad Navaei$^b$, Milad Parvaneh$^b$, Kamran Paynabar$^a$\\[0.5em]
$^a$ Georgia Tech, Atlanta, Georgia, USA \\
$^b$ Ford Motor Company, Dearborn, Michigan, USA
}
\date{}
\maketitle
		 \fi
		
		\if1\blind

            \title{\bf \emph{IISE Transactions} \LaTeX \ Template}
			\author{Author information is purposely removed for double-blind review}
			
\bigskip
			\bigskip
			\bigskip
			\begin{center}
				{\LARGE\bf \emph{IISE Transactions} \LaTeX \ Template}
			\end{center}
			\medskip
            \fi
		\bigskip
		
	\begin{abstract}
    Over the past decades, sensing technologies have advanced rapidly across industries ranging from energy to automotive manufacturing. These systems generate high-dimensional data characterized by complex nonlinear patterns and strong temporal dependencies. Traditional statistical monitoring methods are often limited in their ability to capture such nonlinear structure. Likewise, many analytical approaches used in End-of-Line (EoL) testing rely on predefined thresholds and heuristic rules, which restrict their ability to detect informative anomaly signatures in high-dimensional temporal data. In contrast, while modern deep learning and generative AI models offer strong predictive capabilities, they are often unsuitable in applications where data are costly to collect and where the monitoring system must remain interpretable, low-latency, computationally efficient, and usable by non-technical practitioners. To overcome these limitations, we propose an advanced multivariate monitoring framework for high-dimensional data. The framework operates in two stages. In the first stage, the data are preprocessed to remove incomplete and non-informative samples and to temporally align time series data. In the second stage, nonlinear dimensionality reduction is performed, followed by anomaly detection through a control-chart-based phase I monitoring procedure. The framework can be used in both unsupervised and supervised settings, depending on the availability of ground-truth labels during training. Moreover, its flexible and modular structure allows practitioners to adapt its components to different domains and operational requirements. We evaluate the proposed framework on real production data from an automotive manufacturing environment at Ford Motor Company. The proposed method achieves higher accuracy, recall, and F1 score than the company’s existing model, improving these metrics from 0.50, 0.30, and 0.429 to 0.625, 1.00, and 0.769, respectively.
	\end{abstract}
			
	\noindent%
	{\it Keywords:} Anomaly Detection, Transmission, KPCA, Hotelling's T-squared, Unsupervised Learning.


\section{Introduction}
Failures in transmission systems are particularly consequential because drivetrain related defects constitute a substantial portion of automotive warranty expenditures and long-term reliability concerns. Prior research has demonstrated that warranty databases provide early signals of reliability degradation, highlighting the importance of early detection strategies in automotive systems \cite{zhou2012hazard}. Beyond direct warranty costs, large-scale product defects can trigger recalls that impose significant economic penalties and erode firm value \cite{rhee2006liability}. Empirical evidence in the U.S. automobile industry shows that recalls generate liabilities that negatively affect market performance and consumer perception \cite{rhee2006liability}. Moreover, vehicle defects are associated with measurable safety consequences when crash analyzes indicate that mechanical failures can increase the severity of injury under certain conditions \cite{adanu2024injury}.

Standard End-of-Line (EoL) testing procedures for automotive transmissions are designed to ensure functional correctness before vehicle integration to reduce potential warranty cases. Traditionally, these procedures rely on predefined calibration routines, diagnostic trouble code (DTC) checks, and threshold-based pass/fail criteria executed on dedicated EoL test stations \cite{khond2023development}. In many production environments, vibration and acoustic measurements are used as surrogate indicators of transmission quality, where objective metrics are compared against calibrated limits to determine acceptability \cite{khan2023measurement, bisheh2026stagewise}. Despite these efforts, quality verification has historically depended on relatively simple decision rules, and in some manufacturing settings, destructive teardown inspection of a limited sample is performed to validate batch integrity or investigate suspected defects. However, modern electronically controlled multi-speed transmissions generate vast amounts of high-dimensional (HD) time-series data during EoL testing, including hydraulic pressures, clutch currents, input/output speeds, torque signals, and vibration measurements collected at high sampling rates \cite{liang2025transfer}. 

\begin{figure}[htbp]
    \centering
    \includegraphics[width=1.0\linewidth]{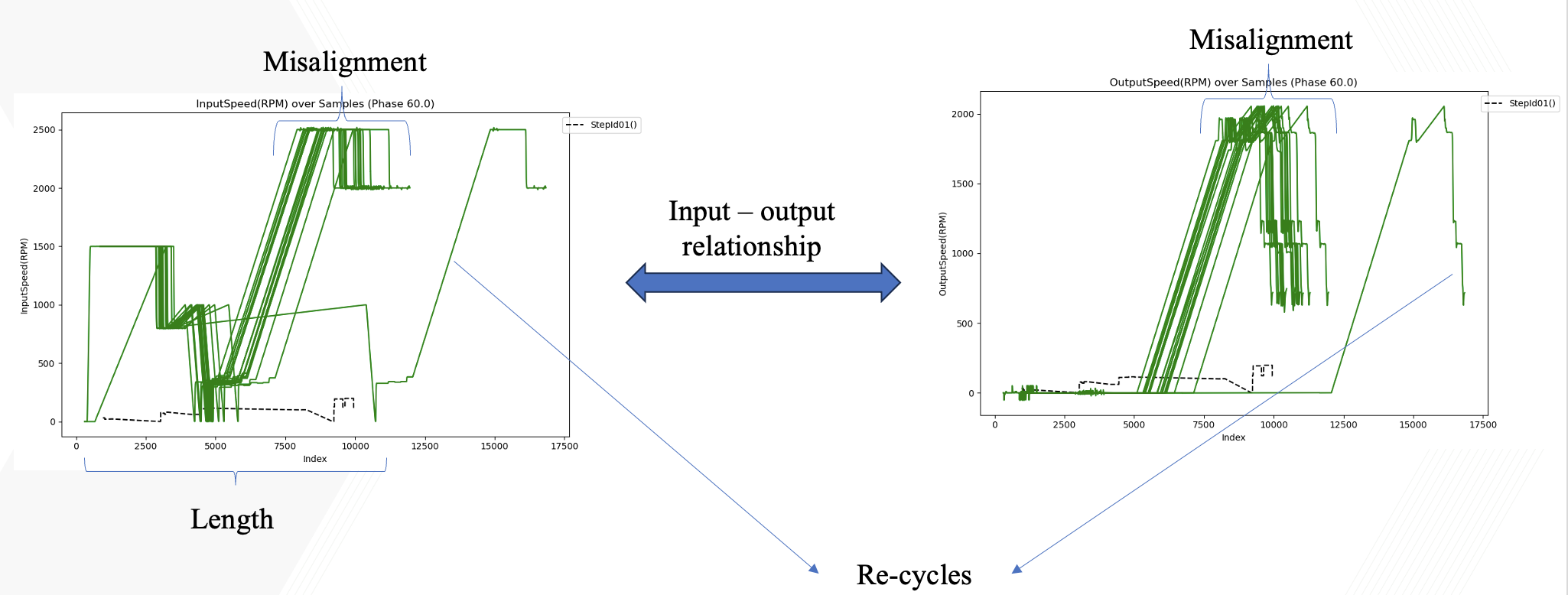}
    \caption{Illustration of key analytical challenges in EoL transmission testing data. The left and right panels show representative input speed and output speed time-series profiles collected during a specific phase. Despite nominally identical test conditions, the signals exhibit temporal misalignment, varying sequence lengths, and repeated cycle segments (re-cycles). In addition, the strong physical coupling between input and output signals highlights the inherent input–output dependency structure that must be considered in data-driven monitoring.}
    \label{fig:intro}
\end{figure}

These data streams present significant analytical challenges. First, the dimensionality of multi-sensor measurements can be extremely large, complicating feature extraction and statistical monitoring. Second, the data exhibit strong temporal dynamics governed by gear shifts, clutch engagements, and transient load responses, making simple static thresholds inadequate. Third, signal misalignment across test cycles and operating phases introduces additional variability that may mask subtle fault signatures. Forth, the test station operators might repeat some tests to ensure test reliability. However, the collected data usually include everything from the starting moment of the test, although only the last cycle  matters. Finally, many transmission signals are inherently coupled through physical input–output relationships, such as torque and speed interactions, leading to multicollinearity and redundancy that can degrade traditional monitoring performance. Collectively, these challenges motivate the development of advanced data-driven methodologies capable of handling HD, temporally complex, and structurally dependent sensor data in EoL transmission testing. These challenges are illustrated in Figure \ref{fig:intro}, where representative input and output speed profiles demonstrate temporal misalignment, variable test lengths, re-cycled segments, and strong input–output dependency, all of which complicate conventional statistical monitoring approaches.

To address the aforementioned challenges in HD and temporally complex EoL transmission data testing, we propose an unsupervised framework for HD profile monitoring and anomaly detection. First, to mitigate signal inconsistency arising from variable sequence lengths, temporal misalignment, and repeated test cycles, we develop a structured preprocessing pipeline that removes recycled segments and applies fast Dynamic Time Warping (DTW) to achieve temporal alignment on a per-signal basis. In addition, to account for the inherent physical coupling between signals, we model input--output relationships via linear regression and utilize residual representations to isolate abnormal behavior from deterministic mechanical dependencies. Then, to capture the nonlinear dynamics governing transmission operation, we employ Kernel Principal Component Analysis (KPCA) to construct a low-dimensional representation of healthy transmission behavior under nominal conditions. This nonlinear embedding preserves complex interactions across pressure, speed, torque, and vibration signals while substantially reducing dimensionality. Building upon this representation, we implement a multivariate statistical monitoring scheme based on Hotelling's $T^2$ statistic to detect deviations from the learned normal profile.

\section{Literature review}

Existing studies on transmission EoL quality control have primarily relied on vibration and acoustic based metrics measured on dedicated test benches, where selected signal features are compared against calibrated thresholds or regression models to assess transmission quality. 
Industrial EoL tester architectures typically employ structured validation logic that integrates diagnostic trouble code verification, parameter consistency checks, and threshold-based decision rules implemented through automated test benches \cite{khond2023development, liang2025transfer}. In transmission applications, vibration and acoustic indicators measured at the EoL stage are often evaluated against calibrated limits or regression-based quality criteria to determine acceptability. \cite{su2024correlation} investigated vibration and acoustic parameters for automatic transmission EoL quality, correlating bench measurements with subjective vehicle evaluation. Related work highlights statistical noise inspections on gear transmissions using multichannel sensor data \cite{horvath2025data} and practical NVH-based EoL anomaly detection prototypes for drive units \cite{blickensdorff2023production,blickensdorffanomaly}. With the increasing availability of high resolution sensor data, recent work has introduced data driven approaches that extract frequency or order domain features from multichannel vibration signals and apply unsupervised anomaly detection algorithms for transmission quality verification \cite{liang2025transfer, jain2023anomaly,giovannardi2024full}. More broadly, machine learning and deep learning methods have been widely investigated for automotive fault diagnosis using multivariate time-series sensor data to improve classification performance and predictive accuracy. Surveys highlight the expanding role of ML-driven predictive maintenance frameworks in vehicle condition monitoring \cite{theissler2021predictive}, and recent studies demonstrate the effectiveness of autoencoder and variational models for deep unsupervised anomaly detection in complex vehicle test environments \cite{hossain2024artificial, geglio2022deep, correia2023ma}.

Data-driven fault detection in rotating machinery has evolved from traditional vibration feature engineering toward advanced machine learning and deep learning frameworks capable of modeling nonlinear and nonstationary dynamics. Early approaches primarily relied on time-domain statistics, frequency-domain analysis, and time--frequency representations to extract fault-sensitive indicators such as RMS, kurtosis, sideband energy, and cyclostationary features for gear and bearing diagnostics \cite{jorani2023gear, peng2019comprehensive, chin2021absolute, shi2022planetary, zhu2025wavelet}. To address varying operating conditions and distribution shifts, recent studies have introduced adaptive time-series models and residual-based strategies, including varying index coefficient autoregression and modified autoregressive formulations for planetary gearboxes operating under speed fluctuations \cite{chen2023modified}. In parallel, deep learning architectures such as CNNs, autoencoders, LSTMs, and hybrid Conv-LSTM networks have been extensively explored to automatically learn spatiotemporal representations from multivariate sensor signals, demonstrating strong classification capability under complex mechanical conditions \cite{wang2025review, shi2022planetary, miao2023application, wei2022two}. Within automotive systems, predictive maintenance frameworks increasingly leverage ensemble learning and multimodal sensor fusion to improve diagnostic robustness across drivetrain and battery subsystems \cite{arena2021predictive, akhtar2026enhancing}. These approaches typically integrate HD sensor streams including vibration, torque, speed, temperature, and electrical signals, aiming to enhance early fault detectability and generalization performance. Despite these advances, challenges remain in handling temporal misalignment, input--output coupling, nonstationarity, and limited labeled failure data, particularly in EoL testing environments where fault signatures may be subtle and operational variability is significant.

HD time-series monitoring in rotating machinery and automotive systems presents significant challenges due to temporal variability, sequence misalignment, and strong coupling among multivariate signals. Under variable operating conditions such as fluctuating rotational speeds and load changes, vibration signatures become nonstationary, leading to distorted spectral features and degraded diagnostic performance \cite{chen2023modified, zhu2025wavelet}. Additionally, varying test durations, cycle repetitions, and speed ramps introduce temporal misalignment and sequence-length variability, complicating direct comparison across samples and limiting the effectiveness of traditional fixed-window feature extraction methods \cite{peng2019comprehensive, chin2021absolute}. Beyond temporal complexity, rotating systems exhibit intrinsic input--output dependencies, where measured vibration responses are strongly influenced by transmission error, structural transfer paths, and dynamic coupling effects \cite{huangfu2023fault, chin2023monitoring}. Such signal coupling can obscure localized fault signatures and induce cross-channel correlations that challenge independent-sensor monitoring strategies. These factors collectively complicate robust anomaly detection in HD monitoring environments, particularly when labeled fault data are scarce and operational variability is substantial.
Dimensionality reduction plays a central role in HD process monitoring, particularly in rotating machinery and automotive systems where multichannel vibration, speed, torque, and transmission error signals generate strongly correlated feature spaces. Classical statistical process monitoring (SPM) frameworks employ latent-variable models such as Principal Component Analysis (PCA) to capture dominant covariance structures and to construct monitoring statistics including Hotelling's $T^2$ and the Squared Prediction Error (SPE) for anomaly detection. In gear crack detection, statistical process control charts such as Shewhart and EWMA have demonstrated effectiveness in identifying deviations from healthy baselines using low-dimensional statistical features \cite{jorani2023gear}. However, linear latent-variable models may be insufficient when system dynamics are nonlinear or operating conditions vary significantly, leading to distorted subspace representations \cite{wang2025tensor}.

To address nonlinearities and nonstationarity in gearbox vibration, recent works have explored advanced time-series modeling and representation learning strategies. For example, varying index coefficient autoregressive models have been proposed to better characterize nonlinear and speed-dependent vibration dynamics before residual-based monitoring \cite{chen2023modified}. In parallel, deep learning approaches such as convolutional and recurrent architectures have been employed to learn compact latent representations from raw sensor streams, effectively performing implicit nonlinear dimensionality reduction prior to fault classification \cite{wang2025review}. In gear wear and transmission error monitoring, model-based and data-driven integration frameworks further emphasize the importance of extracting low-dimensional health indicators that remain stable under operational variability \cite{peng2019comprehensive, chin2023monitoring}. Collectively, these developments highlight the necessity of combining robust dimensionality reduction with multivariate statistical monitoring to achieve reliable detection in complex, HD industrial environments. 

In light of the limitations identified in existing approaches, this study makes several contributions to data-driven monitoring of transmission systems. First, we propose a structured preprocessing framework that addresses inconsistencies in EoL testing data through systematic data cleaning, cycle removal, smoothing, and temporal alignment, enabling reliable comparison across transmissions. Second, we introduce an input--output decoupling step based on regression modeling, where residual signals are used to remove deterministic dependencies between coupled variables and better isolate abnormal behavior. Third, we develop a nonlinear feature extraction approach using KPCA to capture the intrinsic structure of HD multivariate signals in a low-dimensional space. Fourth, we integrate these features into a multivariate statistical monitoring framework based on Hotelling’s $T^2$ statistic for anomaly detection. Fifth, we propose a weighted anomaly scoring and ranking mechanism that aggregates phase-wise deviations to prioritize potentially faulty transmissions. Finally, the proposed framework is validated through a real-world case study using production data from an automotive manufacturing environment, demonstrating its practical applicability and effectiveness under realistic operating conditions.

\section{Methodology}

\begin{figure}[htbp]
    \centering
    \includegraphics[width=0.5\linewidth]{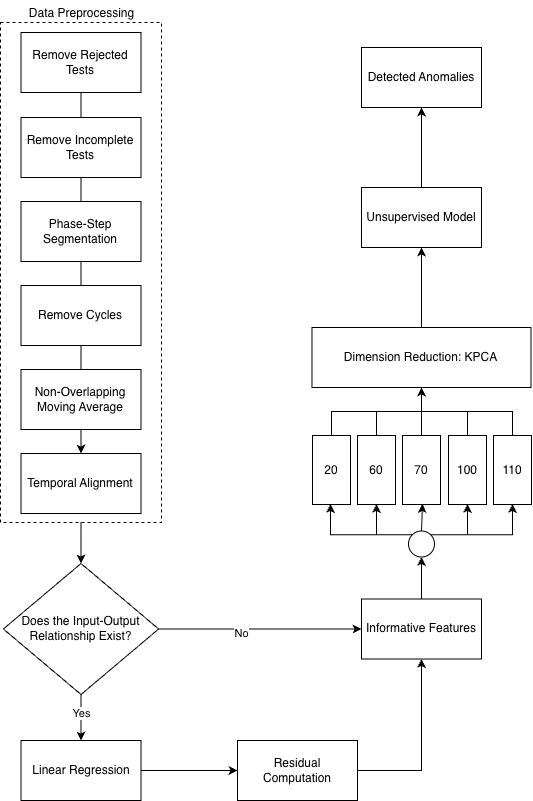}
    \caption{An overview of our methodology: after extracting informative features, dimensionality reduction is applied independently to each phase.}
    \label{fig:method}
\end{figure}

This section presents the proposed methodology, as illustrated in Figure~\ref{fig:method}. The framework begins with a comprehensive preprocessing pipeline designed to ensure data consistency through cleaning, cycle removal, smoothing, and temporal alignment. It then accounts for input--output dependencies via regression-based residual modeling, followed by nonlinear dimensionality reduction using KPCA to extract informative features. Finally, an unsupervised monitoring model is applied to detect potentially anomalous transmissions based on deviations from learned normal behavior. The remainder of this section explains the method in detail.

\subsection{Data Preprocessing}

The raw transmission signals collected from EoL testing are subject to variability due to inconsistent test execution, noise, and repeated operational cycles. To ensure data quality and consistency for subsequent analysis, a structured preprocessing pipeline is applied.

First, transmissions labeled as rejected are removed from the dataset to construct a baseline representing normal operating behavior. Since the proposed framework is unsupervised and designed to learn the characteristics of healthy transmissions, including faulty samples in this stage may bias the learned representation.

Second, incomplete tests are excluded. These correspond to transmissions where required phase--step combinations are missing or improperly recorded, which would otherwise lead to inconsistent feature dimensions and unreliable comparisons across samples.

In addition, repeated cycles within each phase are identified and removed to ensure consistency across transmissions. During EoL testing, certain phases may be executed multiple times due to control adjustments or intermediate resets. These repetitions introduce variability in signal length and distort the representation of normal operation. Therefore, only the final valid cycle is retained for each phase. The procedure for cycle removal is described in Algorithm~\ref{alg:recycle-trim}.

Next, a non-overlapping moving average is applied to each signal to reduce high-frequency noise and stabilize the measurements. Given a time-series signal $x(t)$, the smoothed signal $\tilde{x}(t)$ is obtained by partitioning the data into non-overlapping windows of fixed length $w$ and computing the average within each window. This operation reduces noise while preserving the overall signal trend and also decreases the dimensionality of the data.
\begin{algorithm}
\caption{Removing Recycled Segments by Retaining the Final Cycle}
\label{alg:recycle-trim}
\begin{algorithmic}[1]
\State Initialize empty set $\mathcal{R}$
\For{each phase}
    \State Extract the ordered sequence of step identifiers
    \State Find the minimum distinct step identifier
    \State Find all indices where this minimum identifier appears
    \State Partition these indices into contiguous blocks
    \If{there is only one block}
        \State Append all records of the phase to $\mathcal{R}$
    \Else
        \State Select the last contiguous block
        \State Find the first index of this block
        \State Append records from this index to the end of the phase into $\mathcal{R}$
    \EndIf
\EndFor
\State \Return $\mathcal{R}$
\end{algorithmic}
\end{algorithm}



\subsection{Signal Alignment}

After preprocessing, signals corresponding to the same phase across different transmissions often exhibit variations in temporal progression, where the rate of increase or decrease differs due to uncontrolled testing conditions. These variations lead to temporal misalignment, making direct comparison of time-series signals unreliable. To address this issue, we employ DTW \cite{salvador2007toward_fastdtw} to align signals and remove distortions caused by inconsistent execution speeds.

DTW is a widely used technique for measuring similarity between time series with temporal distortions by allowing nonlinear alignment along the time axis. Given two time series 
\[
X = (x_1, x_2, \dots, x_N), \quad Y = (y_1, y_2, \dots, y_M),
\]
DTW seeks an optimal warping path
\[
W = (w_1, w_2, \dots, w_L), \quad w_l = (i_l, j_l),
\]
which aligns elements $x_{i_l}$ and $y_{j_l}$ while satisfying the following constraints:
\begin{itemize}
    \item \textbf{Boundary condition:} $w_1 = (1,1)$ and $w_L = (N,M)$
    \item \textbf{Monotonicity:} $i_{l+1} \geq i_l$, $j_{l+1} \geq j_l$
    \item \textbf{Continuity:} $i_{l+1} - i_l \leq 1$, $j_{l+1} - j_l \leq 1$
\end{itemize}

The optimal warping path minimizes the cumulative alignment cost:
\[
\mathrm{Dist}(W) = \sum_{l=1}^{L} d(x_{i_l}, y_{j_l}),
\]
where $d(\cdot,\cdot)$ is typically the squared Euclidean distance. This optimization is solved via dynamic programming using the recursive formulation:
\[
D(i,j) = d(x_i, y_j) + \min \{ D(i-1,j),\; D(i,j-1),\; D(i-1,j-1) \},
\]
where $D(i,j)$ represents the minimum cumulative cost up to indices $(i,j)$.

The classical DTW algorithm has a time and space complexity of $O(NM)$, which becomes computationally prohibitive for long time series. To improve efficiency, we adopt the FastDTW algorithm \cite{salvador2007toward_fastdtw}, which approximates the optimal warping path with linear time and space complexity. FastDTW employs a multilevel approach that recursively computes alignments at coarser resolutions and refines them at higher resolutions, significantly reducing computational cost while maintaining high alignment accuracy. Using FastDTW, each signal is temporally aligned to a reference trajectory, ensuring consistent representation across transmissions. This alignment step enables meaningful comparison of signals and facilitates subsequent feature extraction and monitoring.

\subsection{Input--Output Relationship Modeling}

In transmission systems, many sensor signals are not independent but are physically coupled through underlying mechanical relationships. For example, output torque is directly influenced by input torque through gear interactions and load transfer mechanisms. As a result, variations observed in one signal may not necessarily indicate abnormal behavior if they are consistent with corresponding changes in its paired signal. Relying solely on raw measurements can therefore lead to misleading interpretations, where normal operational dependencies are incorrectly identified as anomalies.

To address this issue, we explicitly model the input--output relationship between paired signals and remove the deterministic dependency component. Let $x(t)$ denote an input signal (e.g., input torque) and $y(t)$ denote the corresponding output signal (e.g., output torque). We assume a functional relationship of the form
\[
y(t) = f(x(t)) + \epsilon(t),
\]
where $f(\cdot)$ captures the systematic dependency and $\epsilon(t)$ represents deviations from this relationship.

In this work, we approximate $f(\cdot)$ using a linear regression model:
\[
y(t) = \beta_0 + \beta_1 x(t) + \epsilon(t),
\]
where $\beta_0$ and $\beta_1$ are regression coefficients estimated from the data. The residual term is then computed as
\[
r(t) = y(t) - \hat{y}(t) = y(t) - (\hat{\beta}_0 + \hat{\beta}_1 x(t)).
\]

The residual signal $r(t)$ represents the portion of the output that cannot be explained by the input--output dependency. By construction, it removes the dominant linear coupling between signals and isolates deviations that are more likely to correspond to abnormal behavior.

This transformation provides several advantages for anomaly detection. First, it reduces spurious variability caused by normal operational dependencies, enabling more meaningful comparisons across transmissions. Second, it produces features that are approximately invariant to changes in operating conditions that affect both signals simultaneously. Finally, it enhances the sensitivity of downstream monitoring methods by focusing on deviations from expected behavior rather than absolute signal magnitudes. The resulting residual signals are therefore used as inputs for subsequent feature extraction and monitoring.

\subsection{Low-Dimensional Representation via Kernel PCA}

The preprocessed signals consist of HD, multivariate time-series data collected across multiple sensors and operational segments. Such data are typically characterized by strong correlations, redundancy, and noise, which can degrade the performance of monitoring algorithms due to the curse of dimensionality and instability in covariance estimation. Therefore, it is necessary to extract a compact representation that preserves the essential structure of normal system behavior while reducing noise and redundancy.

Classical dimensionality reduction techniques such as Principal Component Analysis (PCA) project data onto a linear subspace that captures the maximum variance. However, in transmission systems, the underlying relationships among signals are often nonlinear due to complex mechanical interactions, load variations, and dynamic coupling effects. As a result, linear PCA may fail to capture the intrinsic structure of the data. To address this limitation, we employ KPCA, which extends PCA to nonlinear feature spaces through the use of kernel functions.

Let $\mathbf{x}_1, \mathbf{x}_2, \dots, \mathbf{x}_n \in \mathbb{R}^d$ denote the set of input feature vectors obtained after preprocessing and residual computation. KPCA implicitly maps the data into a HD feature space $\mathcal{F}$ via a nonlinear mapping $\phi(\cdot)$ and performs PCA in that space. Instead of explicitly computing $\phi(\mathbf{x})$, KPCA relies on a kernel function defined as
\[
K(\mathbf{x}_i, \mathbf{x}_j) = \langle \phi(\mathbf{x}_i), \phi(\mathbf{x}_j) \rangle.
\]

In this work, we use a polynomial kernel of the form
\[
K(\mathbf{x}_i, \mathbf{x}_j) = (a \, \mathbf{x}_i^\top \mathbf{x}_j + b)^p,
\]
where $a$, $b$, and $p$ are kernel parameters controlling the nonlinearity. A polynomial kernel was selected because it can capture nonlinear interactions among transmission signals while retaining a structured, interpretable similarity measure based on inner products. The kernel matrix $\mathbf{K} \in \mathbb{R}^{n \times n}$ is first centered as
\[
\tilde{\mathbf{K}} = \mathbf{K} - \mathbf{1}_n \mathbf{K} - \mathbf{K} \mathbf{1}_n + \mathbf{1}_n \mathbf{K} \mathbf{1}_n,
\]
where $\mathbf{1}_n$ is an $n \times n$ matrix with all entries equal to $1/n$.

KPCA then solves the eigenvalue problem
\[
\tilde{\mathbf{K}} \boldsymbol{\alpha} = \lambda \boldsymbol{\alpha},
\]
where $\lambda$ and $\boldsymbol{\alpha}$ denote the eigenvalues and eigenvectors of the centered kernel matrix. The projection of a new sample $\mathbf{x}$ onto the $k$-th principal component is given by
\[
z_k(\mathbf{x}) = \sum_{i=1}^{n} \alpha_i^{(k)} K(\mathbf{x}_i, \mathbf{x}).
\]

The resulting low-dimensional representation $\mathbf{z} = (z_1, z_2, \dots, z_q)$ captures the dominant nonlinear structure of the data while filtering out noise and redundant variations. This representation provides a compact and informative feature space for subsequent multivariate monitoring and anomaly detection.

\subsection{Monitoring Statistics and Anomaly Detection}

After obtaining the low-dimensional representation from KPCA, each transmission is represented by a vector of principal component scores. These score vectors provide a compact feature space in which the normal operating behavior can be statistically characterized and monitored. Since each phase contains a different number of retained features, monitoring is performed separately for each phase.

Let $\{\mathbf{z}_i\}_{i=1}^m$ denote the in-control score vectors for a given phase, where $\mathbf{z}_i \in \mathbb{R}^p$ and $p$ is the feature dimension of that phase. The in-control mean vector is estimated as
\[
\boldsymbol{\mu} = \frac{1}{m}\sum_{i=1}^m \mathbf{z}_i,
\]
and the sample covariance matrix is estimated by
\[
\boldsymbol{\Sigma} = \frac{1}{m-1}\sum_{i=1}^m (\mathbf{z}_i-\boldsymbol{\mu})(\mathbf{z}_i-\boldsymbol{\mu})^\top,
\]
where $\boldsymbol{\Sigma}\in\mathbb{R}^{p\times p}$.

For a new observation $\mathbf{z}$, the Hotelling's $T^2$ statistic is computed as
\[
T^2 = (\mathbf{z}-\boldsymbol{\mu})^\top \boldsymbol{\Sigma}^{-1} (\mathbf{z}-\boldsymbol{\mu}).
\]
This statistic measures the squared Mahalanobis distance from the in-control center in the reduced feature space. Larger values of $T^2$ indicate greater deviation from normal behavior.

For individual observations in Phase I monitoring, the control limits are determined using the Beta-distribution-based formulation:
\begin{equation}
\mathrm{UCL} = \frac{(m-1)^2}{m}\,\beta_{\alpha,\;p/2,\;(m-p-1)/2},
\qquad
\mathrm{LCL} = 0,
\end{equation}
where $\beta_{\alpha,\;p/2,\;(m-p-1)/2}$ denotes the upper $\alpha$ percentile of a Beta distribution with parameters $p/2$ and $(m-p-1)/2$ \cite{montgomery2009isqc_montgomery}. An observation is flagged as anomalous if its $T^2$ value exceeds the upper control limit.

Because multiple phases are monitored separately, we allocate the overall Type I error across phases. Let $\gamma$ denote the desired overall false alarm probability and let $\alpha$ denote the per-phase significance level. Assuming the five phase-specific control charts are assigned the same Type I error, the overall false alarm probability satisfies
\[
\gamma = 1-(1-\alpha)^5.
\]
Solving for $\alpha$ gives
\[
\alpha = 1-(1-\gamma)^{1/5}.
\]

By applying Hotelling's $T^2$ statistic in the KPCA score space, the proposed framework combines nonlinear feature extraction with multivariate statistical process monitoring, enabling systematic detection of deviations from normal transmission behavior.

\subsection{Anomaly Score Computation}
The methodology we described before would give us a list of potentially anomalous serial numbers for each phase. One can behave very conservative and investigate the union of the five sets as the transmissions that are worth further investigation. However this method would make sure that no potentially anomalous transmission would pass to deployment without further investigation, the conservative nature of it will lead to a large number of false positives. 

To gain a deeper understanding of the anomaly magnitude, we define an anomaly score in Alg.~\ref{alg:ranking}. This algorithm computes the anomaly score as the weighted summation of anomaly scores across different phases. To compute the anomaly score for each phase, we first calculate the absolute difference between the statistic and its corresponding threshold. For each phase, the difference is divided by the threshold so anomaly score would reflect the ratio of deviation compared to the control limit or threshold. This would prevent a phase with larger threshold to dominate the anomaly score. When using Hotelling's T-squared, the threshold would be the upper control limit computed from the Beta distribution. 

In addition to Phase I Hotelling’s $T^2$ for anomaly detection, any one-class classification method can also be employed. In this case, the anomaly score is computed relative to a threshold defined at the $(1-\alpha)$ quartile, where $\alpha$ denotes the contamination rate.

To choose the weights corresponding to each phase, if no prior information is available one can consider all of them to be equal. Otherwise, expert input can guide the weight selection. Later in this paper, we will discuss a feedback loop from the deployment in cars and the warranty feedback after some time and discuss the way to improve our model.

\begin{algorithm}[htbp]
\caption{Ranking Samples Based on Weighted Anomaly Score}
\label{alg:ranking}
\begin{algorithmic}[1]
\State \textbf{Input:} Set of phases $\mathcal{L}$, sample $s$ phase specific anomaly score $a_{s,\ell}$, thresholds $\{t_{\ell}\}_{\ell \in \mathcal{L}}$, weights $\{w_{\ell}\}_{\ell \in \mathcal{L}}$ such that $\sum_{\ell \in \mathcal{L}} w_{\ell} = 1$
\State \textbf{Input:} Set of samples $\mathcal{S}$
\For{each sample $s \in \mathcal{S}$}
    \State Initialize anomaly score $A(s) \gets 0$
    \For{each phase $\ell \in \mathcal{L}$}
        \State Define indicator:
        \[
        I_{\ell}(s) =
        \begin{cases}
        1, & \text{if $s$ is out-of-control in phase $\ell$} \\
        0, & \text{otherwise}
        \end{cases}
        \]
        \State Update anomaly score:
        \[
        A(s) \gets A(s) + w_{\ell} \cdot I_{\ell}(s) \cdot \frac{\big(a_{s,\ell} - t_{\ell}\big)}{t_{\ell}}
        \]
    \EndFor
\EndFor
\State Sort all samples $s \in \mathcal{S}$ in descending order of $A(s)$
\State \textbf{Output:} Ranked list of samples
\end{algorithmic}
\end{algorithm}


\section{Real Case Study Data Description}

\begin{figure}[!htbp]
    \centering
    \includegraphics[width=0.9\linewidth]{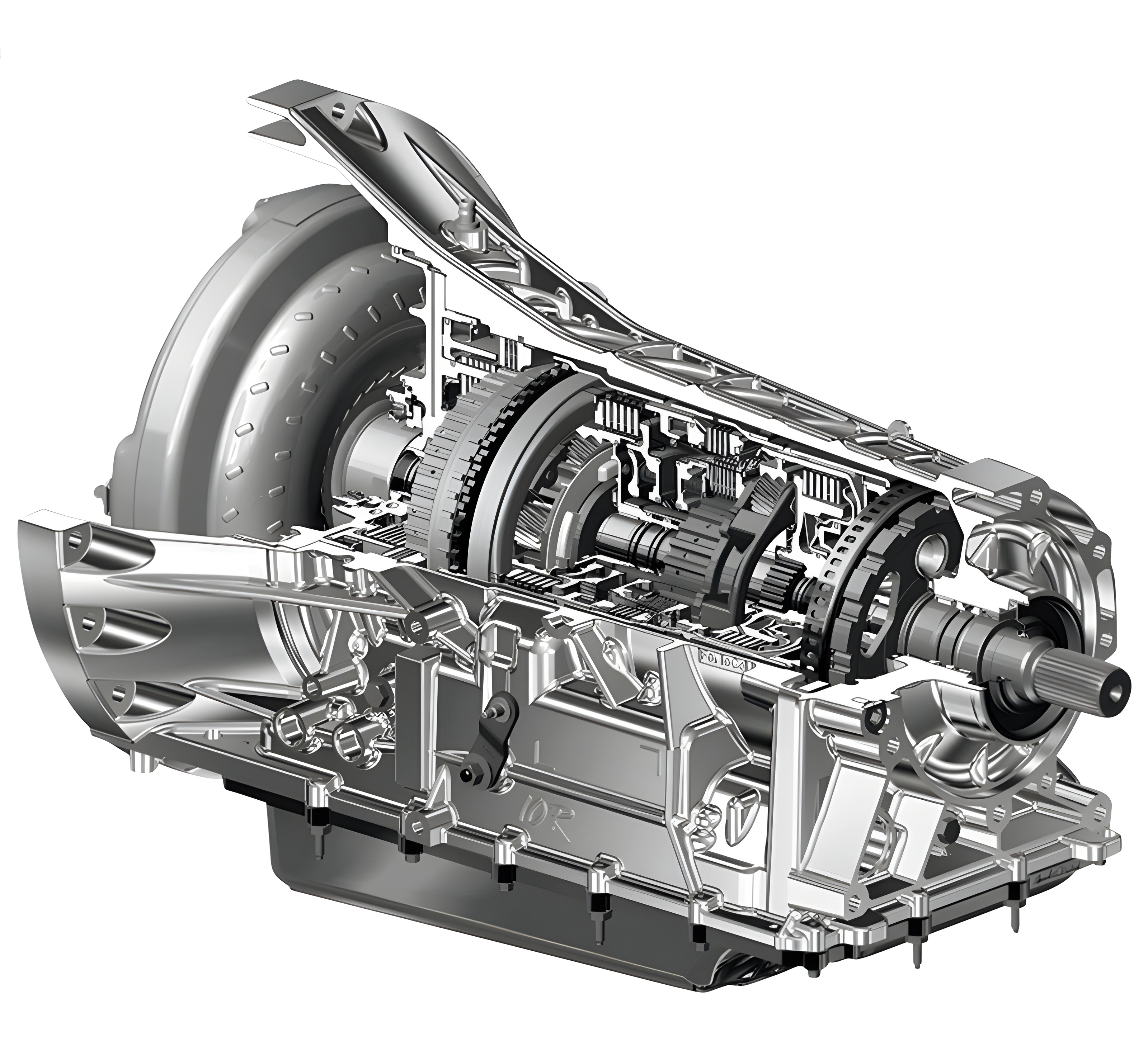}
    \caption{Ford 10R80 ten-speed automatic transmission(~\cite{ford10r80}).}
    \label{fig:ford10r80}
\end{figure}

The dataset used in this study consists of EoL testing data collected from 502 transmission units over a three-day period spanning September 30, October 1, and October 2, 2024. An image of the transmission is shown in Figure~\ref{fig:ford10r80}. For each transmission, multiple sensor signals were recorded during standardized testing procedures designed to evaluate the operational performance of the system.

The recorded data includes multivariate time-series measurements such as torque, speed, and other relevant process variables, collected across different operational phases of the testing process. Each transmission is therefore represented by a set of HD signals corresponding to multiple sensors and testing segments. Due to variations in test execution, the data exhibit differences in sequence length, temporal progression, and signal behavior across transmissions.

The recorded data were collected across multiple phases, with several steps defined within each phase. We refer to the phases using numerical identifiers and to the steps using chronological identifiers. Due to data confidentiality, we are unable to disclose the specific meaning of each phase or step; however, these phase–step definitions were established by domain experts at the company.

As is typical in industrial EoL testing environments, the dataset contains inherent variability arising from measurement noise, operational fluctuations, and system dependencies among signals. 

\begin{figure}[!htbp]
    \centering
    \includegraphics[width=0.9\linewidth]{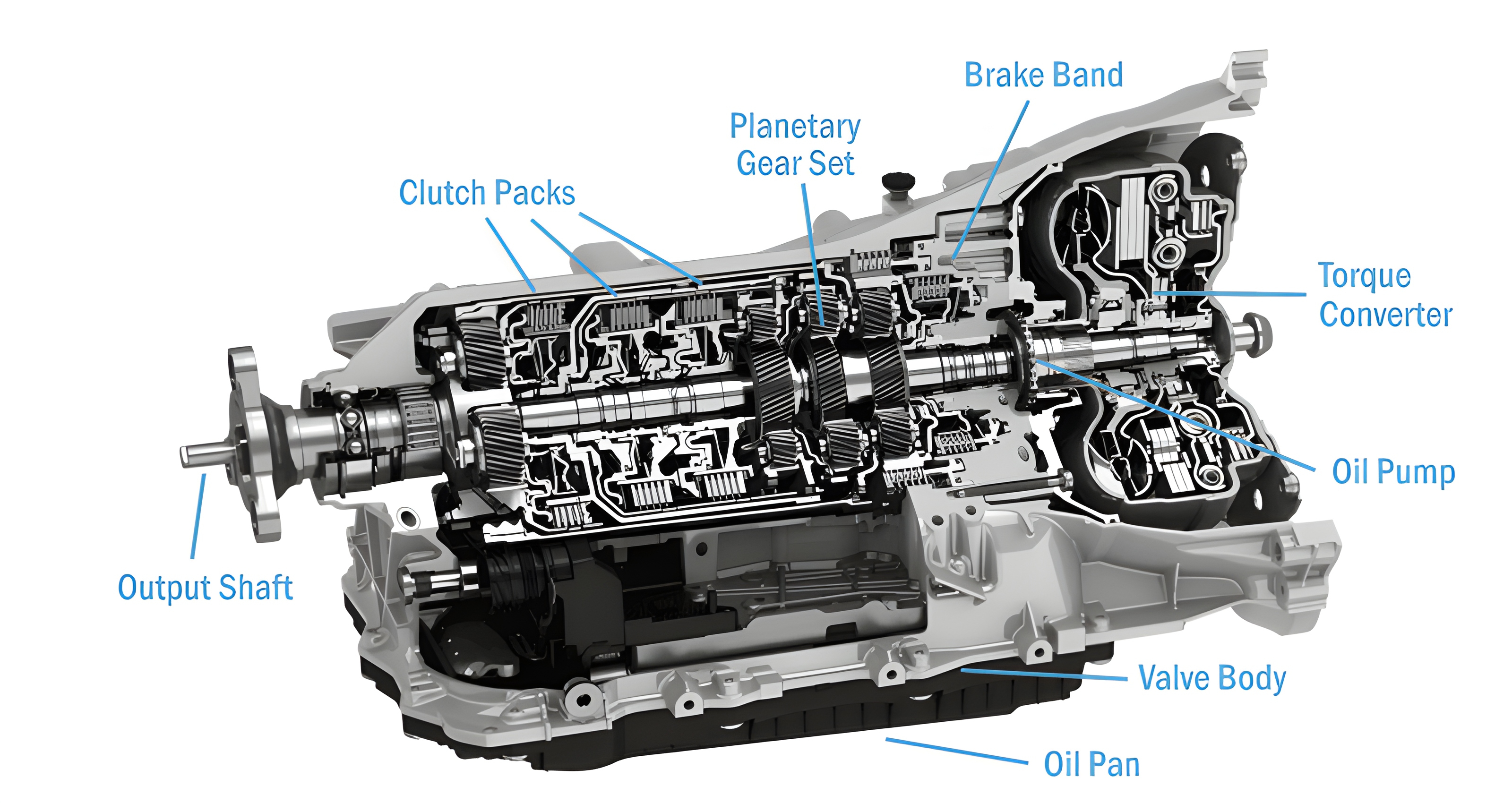}
    \caption{Illustrative diagram of an automatic transmission with labeled components(~\cite{engchoice}).}
    \label{fig:transmission_guide}
\end{figure}

The dataset includes both in-control and out-of-control transmissions, with the latter identified based on quality inspection outcomes. This structure enables the evaluation of the proposed unsupervised monitoring framework in detecting anomalous behavior under realistic manufacturing conditions. The data includes the features shown in Tab.~\ref{tab:signals_description}. A visual description of the transmission components is shown in Figure~\ref{fig:transmission_guide}.
\begin{longtable}{%
  >{\raggedright\arraybackslash}p{3.2cm}%
  >{\raggedright\arraybackslash}p{4.2cm}%
  >{\raggedright\arraybackslash}p{1.8cm}%
  >{\raggedright\arraybackslash}p{6.2cm}}
\caption{Summary of transmission signals grouped by category. This table was generated by the help of Google Gemini.}\label{tab:signals_description}\\
\toprule
\thead{Category} & \thead{Signal Name} & \thead{Units} & \thead{Description} \\
\midrule
\endfirsthead

\toprule
\thead{Category} & \thead{Signal Name} & \thead{Units} & \thead{Description} \\
\midrule
\endhead

\bottomrule
\endfoot

\makecell[l]{Pressure\\Signals} & Clutch Pressure Command (A--F) & PSI & Commanded pressure for clutch packs A--F, used to engage gears. \\
\makecell[l]{Pressure\\Signals} & Torque Converter Outlet Pressure & PSI & Fluid pressure at the outlet of the torque converter. \\
\makecell[l]{Pressure\\Signals} & Main Line Pressure & PSI & Actual main hydraulic line pressure in the valve body. \\
\makecell[l]{Pressure\\Signals} & Main Line Pressure Command & PSI & Commanded main line pressure requested by the transmission control module. \\
\makecell[l]{Pressure\\Signals} & Filtered Line Pressure & PSI & Smoothed version of line pressure for stable control. \\
\makecell[l]{Pressure\\Signals} & Line Pressure Modulation Signal & PSI & High-frequency pressure modulation used for solenoid cleaning or fine clutch control. \\

\midrule
\makecell[l]{Rotational\\Speed Signals} & Input Shaft Speed & RPM & Rotational speed of the input shaft, typically driven by the engine. \\
\makecell[l]{Rotational\\Speed Signals} & Redundant Input Shaft Speeds & RPM & Two input shaft speed sensor measurements used for redundancy or monitoring. \\
\makecell[l]{Rotational\\Speed Signals} & Output Shaft Speed & RPM & Rotational speed of the output shaft driving the wheels. \\
\makecell[l]{Rotational\\Speed Signals} & Output Shaft Sensor Frequency & Hz & Raw frequency signal from the output shaft speed sensor. \\
\makecell[l]{Rotational\\Speed Signals} & Turbine Shaft Speed & RPM & Turbine shaft speed inside the torque converter, used to monitor slippage. \\

\midrule
\makecell[l]{Torque\\Signals} & Transmission Input Torque & lb-ft & Estimated torque delivered from the engine to the transmission. \\
\makecell[l]{Torque\\Signals} & Transmission Output Torque & lb-ft & Estimated torque delivered from the transmission to the driveshaft. \\

\midrule
\makecell[l]{Electrical \&\\Solenoid Signals} & Main Line Solenoid Command Current & mA & Commanded current to the main line pressure solenoid. \\
\makecell[l]{Electrical \&\\Solenoid Signals} & Torque Converter Clutch Command Current & mA & Commanded current for the torque converter clutch solenoid. \\
\makecell[l]{Electrical \&\\Solenoid Signals} & Individual Solenoid Currents & mA & Measured currents for individual solenoids. \\
\makecell[l]{Electrical \&\\Solenoid Signals} & Main Line Solenoid Current & mA & Measured current at the main line pressure solenoid. \\
\makecell[l]{Electrical \&\\Solenoid Signals} & Torque Converter Clutch Solenoid Current & mA & Measured current at the torque converter clutch solenoid. \\
\makecell[l]{Electrical \&\\Solenoid Signals} & Range Sensor Duty Cycle & \% & Duty cycle of the transmission range sensor, indicating gear selector position. \\

\midrule
\makecell[l]{Test \&\\State Signals} & Test Phase / Step Identifier & -- & Identifiers for the current test phase and step. \\
\makecell[l]{Test \&\\State Signals} & Test Rejection Flag & Binary & Flag indicating whether the transmission test was rejected or failed. \\
\end{longtable}

\begin{table}[htbp]
\centering
\caption{Summary of phases, their steps, and associated important signals.}
\label{tab:phase_steps}
\renewcommand{\arraystretch}{1.3}
\begin{tabularx}{\linewidth}{>{\raggedright\arraybackslash}p{2cm} 
                                >{\raggedright\arraybackslash}X 
                                >{\raggedright\arraybackslash}X}
\toprule
\textbf{Phase} & \textbf{Steps} & \textbf{Important Signals} \\
\midrule
Phase 20 & [20, 22, 32] & ['Range Sensor Duty Cycle (\%)', 'Transmission Output Torque (lb-ft)', 'Solenoid 2 Current (mA)', 'Solenoid 4 Current (mA)'] \\
\midrule
Phase 60 & [1, 61, 71, 77, 80] & ['Residual of Input Shaft Speed and Output Shaft Speed (RPM)', 'Residual of Input Torque and Output Torque (lb-ft)'] \\
\midrule
Phase 70 & [61, 110, 111, 112, 113, 114, 115] & ['Input Shaft Speed (RPM)', 'Filtered Line Pressure (PSI)', 'Main Line Pressure (PSI)'] \\
\midrule
Phase 100 & [100, 101] & ['Input Shaft Speed (RPM)', 'Transmission Input Torque (lb-ft)', 'Turbine Shaft Speed (RPM)', 'Torque Converter Outlet Pressure (PSI)'] \\
\midrule
Phase 110 & [1, 121, 123, 126, 193, 198] & ['Residual of Input Shaft Speed and Output Shaft Speed (RPM)', 'Residual of Input Torque and Output Torque (lb-ft)', 'Residual of Turbine Shaft Speed and Output Shaft Speed (RPM)'] \\
\bottomrule
\end{tabularx}
\end{table}

\section{Real Case Study Implementation Results}

Here in this section we implemented our proposed methodology step by step on the dataset and reported the result.

\subsection{Dropping Rejected Test}
The main aim of our algorithm to estimate the distribution of in-control transmission sensor and use that to find anomalous transmissions. To this end, we drop all transmissions that have been rejected. In addition to rejected tests which their labels exist, some tests are incomplete. This manifests itself as a test does not include all required (phase id, step id) shown in Tab.~\ref{tab:phase_steps}. These phase–step pairs correspond to critical segments identified by company domain experts as likely to exhibit transmission signatures. A summary of the total number of transmissions tested, total rejections and total incomplete tests is shown in Tab.~\ref{tab:dataset_summary}.

\begin{table}[ht]
\centering
\resizebox{\textwidth}{!}{%
\begin{tabular}{lcccc}
\hline
\textbf{Dataset} & \textbf{Total Transmissions} & \textbf{Rejected Test} & \textbf{Complete Test} & \textbf{Not Complete Test} \\
\hline
Data from 2025  & 502 & 12 & 476 & 26 \\
\hline
\end{tabular}
}
\caption{Summary of transmission dataset.}
\label{tab:dataset_summary}
\end{table}

\subsection{Removing Cycles}
\begin{figure}[!htbp]
\centering
\includegraphics[width=\textwidth]{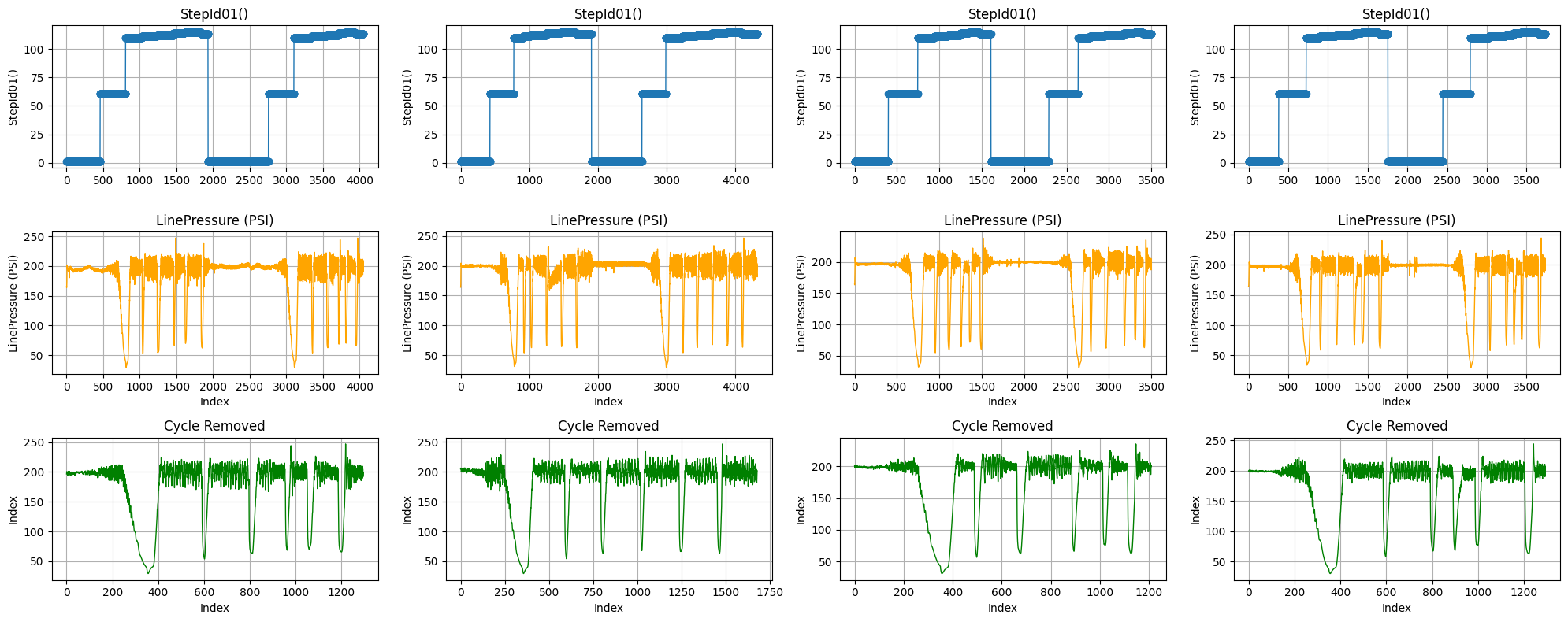}
\caption{Line pressure(PSI) in phase 70.00. Each column shows a distinct transmission, first row is the steps in phase 70, second row is line pressure(PSI), and third row is the same after removing cycles.}
\label{Fig:recycles}
\end{figure}

If a recycle happens for any reason, only the last cycle would be acceptable. To detect if a recycle happens or not, for each phase, we iterate through it and find the minimum step id value present in it. Then we check if this step id is present in different non-contagious blocks. If so, we trim everything but the last cycle. 
In Figure~\ref{Fig:recycles} some examples of before and after recycle removing are shown. 

\begin{table}[!htbp]
\centering
\caption{Recycle detection summary for the transmission dataset. 
For each phase, values are shown as the ratio of recycled to the transmissions containing that phase (\%).}
\label{tab:recycle_summary}
\resizebox{\textwidth}{!}{%
\begin{tabular}{lcccccc}
\toprule
Total Transmissions & Transmissions w/ Recycle & Phase 20 & Phase 60 & Phase 70 & Phase 100 & Phase 110 \\
\midrule
 502 & 8 &
0 / 481 (0.0\%) & 4 / 480 (0.8\%) & 4 / 478 (0.8\%) & 0 / 477 (0.0\%) & 0 / 476 (0.0\%) \\
\bottomrule
\end{tabular}%
}
\end{table}

\subsection{Incomplete Test Detection}
After removing cycles within each phase, some transmissions become incomplete because they no longer contain the required phase–step pairs listed in Table~\ref{tab:phase_steps}. We identify four such transmissions and exclude them from further analysis. As a result, 472 out of 502 transmissions are retained for the subsequent steps. For each retained transmission, we include only the data corresponding to the phase–step pairs specified in Table~\ref{tab:phase_steps}. 
\subsection{Non-Overlapping Moving Average}
In this step, a non-overlapping moving average is applied with window size equal to 3 to make the effective length smaller by a factor of 3.
\subsection{Signal Alignments}
After applying non overlapping moving average, each sensor across each transmission shows different rate in which the measurement increases or decreases. This can lead to differences in sensor measurements that cannot be explained by the transmission behavior itself but by some unknown condition during the test time. To remove the ambient condition effect and register the sensors readings, we use DTW. 

\begin{figure}[!htbp]
\centering
\setlength{\tabcolsep}{6pt}
\begin{tabular}{@{}p{.48\linewidth}p{.48\linewidth}@{}}
\textbf{Before applying DTW} & \textbf{After applying DTW} \\[0.0em]
\includegraphics[width=0.9\linewidth]{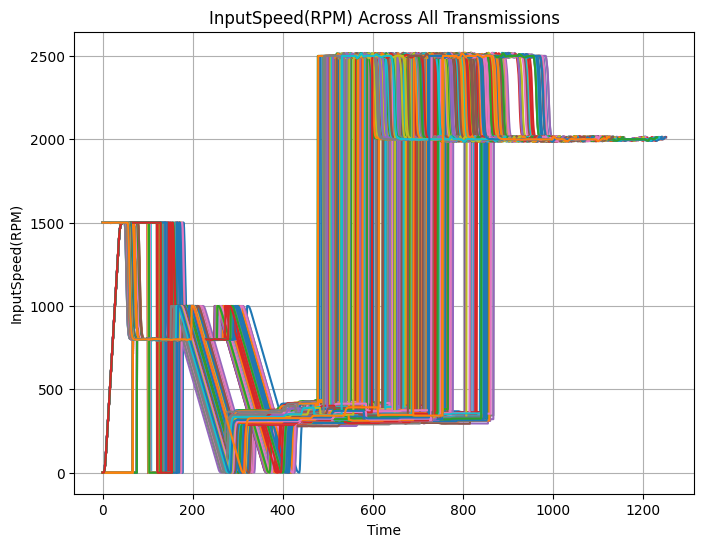} &
\includegraphics[width=0.9\linewidth]{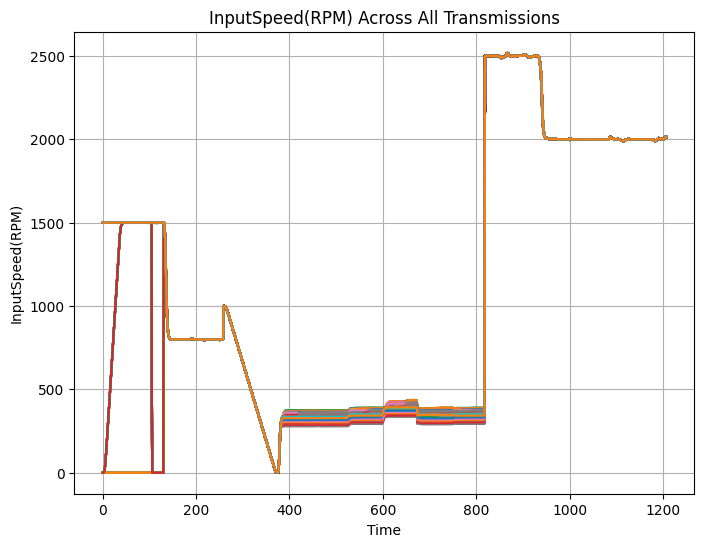} \\[0.0em]

\includegraphics[width=0.9\linewidth]{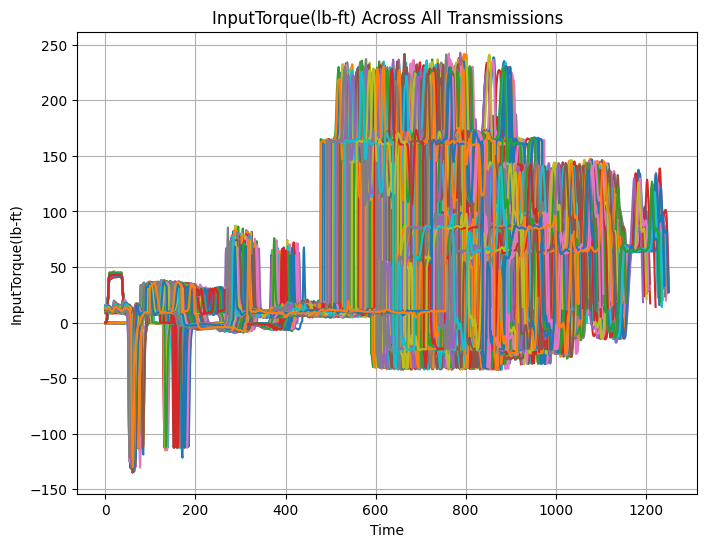} &
\includegraphics[width=0.9\linewidth]{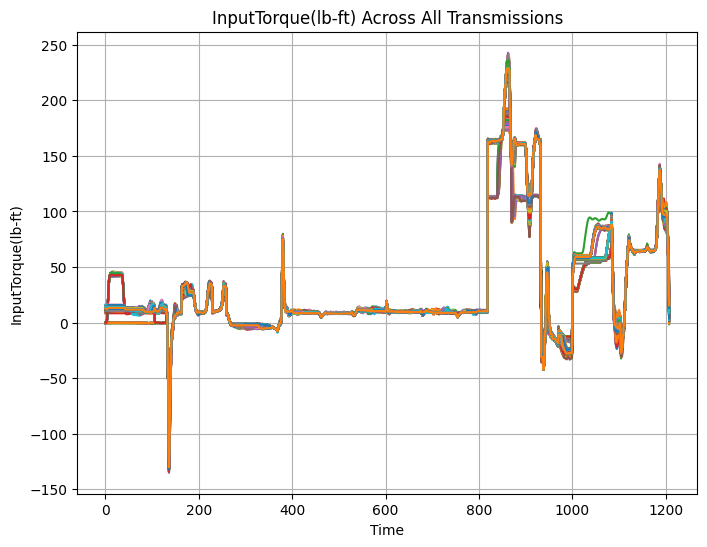} \\[0.0em]

\includegraphics[width=0.9\linewidth]{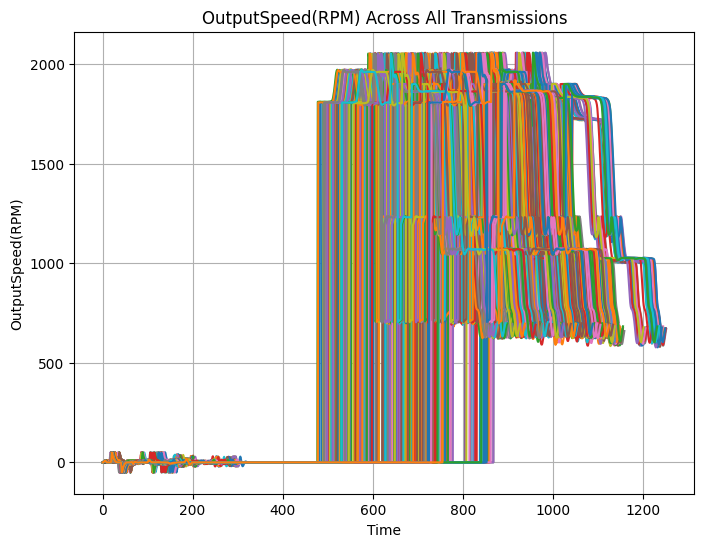} &
\includegraphics[width=0.9\linewidth]{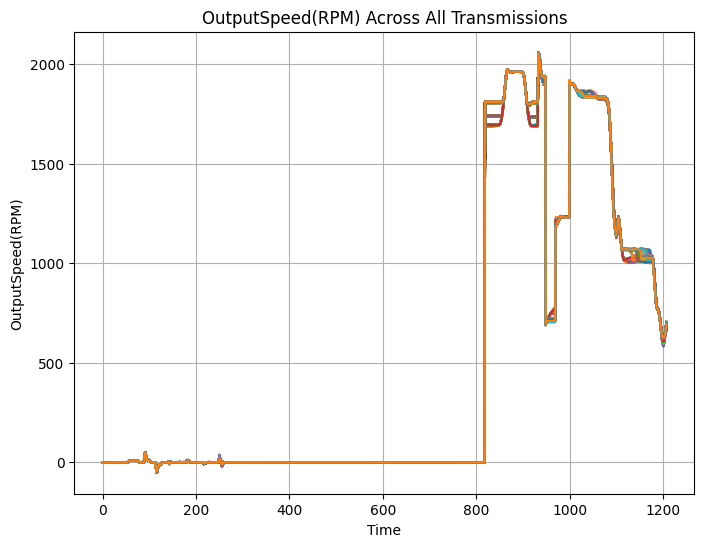} \\[-1em]


\includegraphics[width=0.9\linewidth]{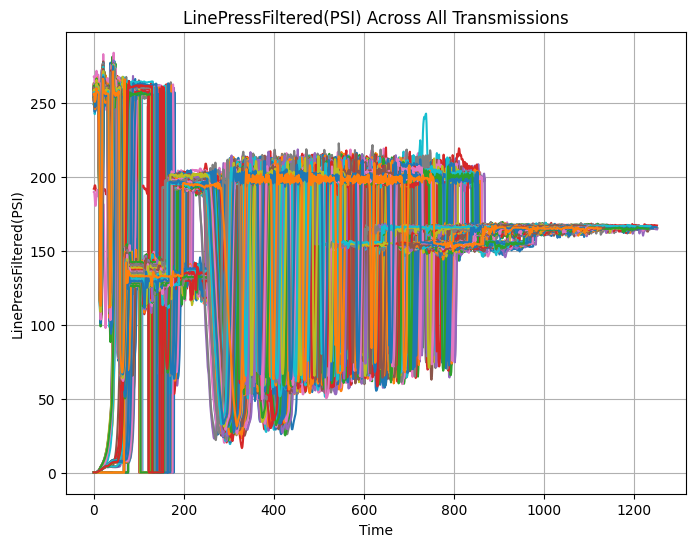} &
\includegraphics[width=0.9\linewidth]{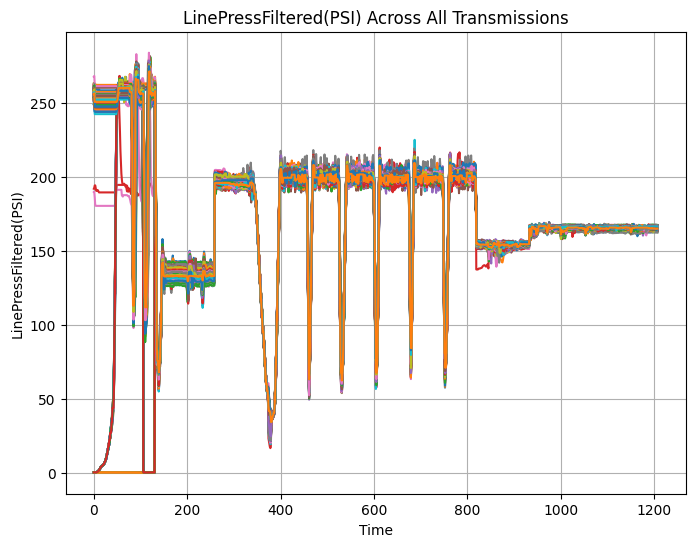} \\
\end{tabular}
\caption{Signal traces (Part 1): before vs.\ after DTW.}
\label{fig:dtw-part1}
\end{figure}

\begin{figure}[!htbp]
\centering
\setlength{\tabcolsep}{6pt}
\begin{tabular}{@{}p{.48\linewidth}p{.48\linewidth}@{}}
\textbf{Before applying DTW} & \textbf{After applying DTW} \\[0.0em]
\includegraphics[width=0.9\linewidth]{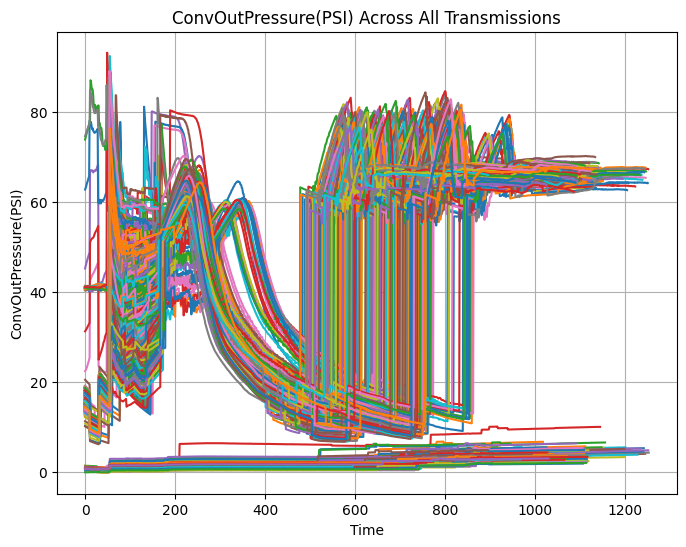} &
\includegraphics[width=0.9\linewidth]{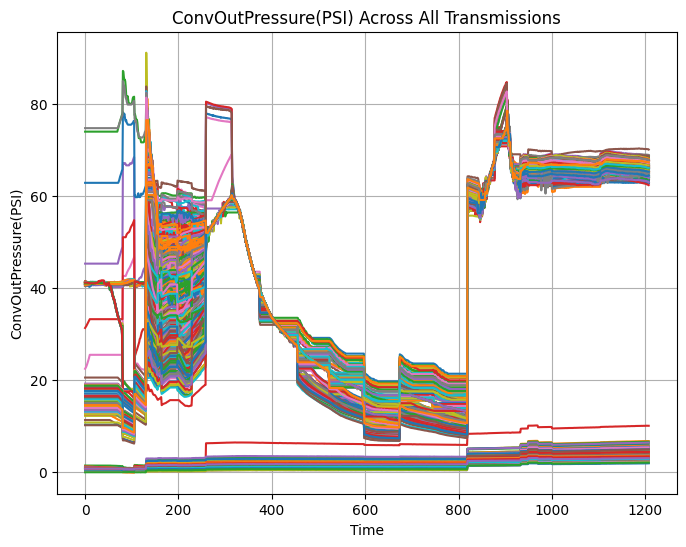} \\[0.0em]
\includegraphics[width=0.9\linewidth]{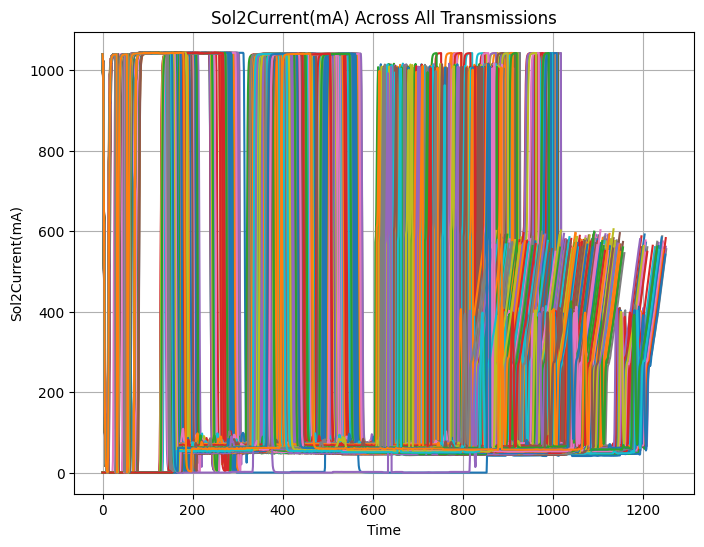} &
\includegraphics[width=0.9\linewidth]{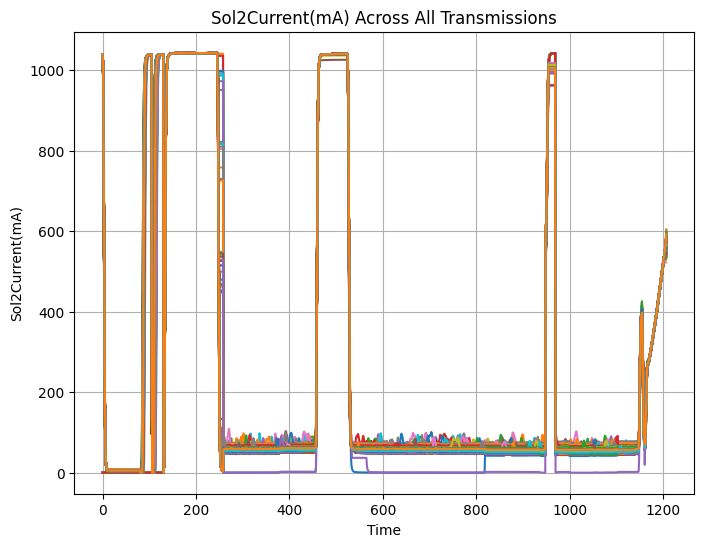} \\[0.0em]
\includegraphics[width=0.9\linewidth]{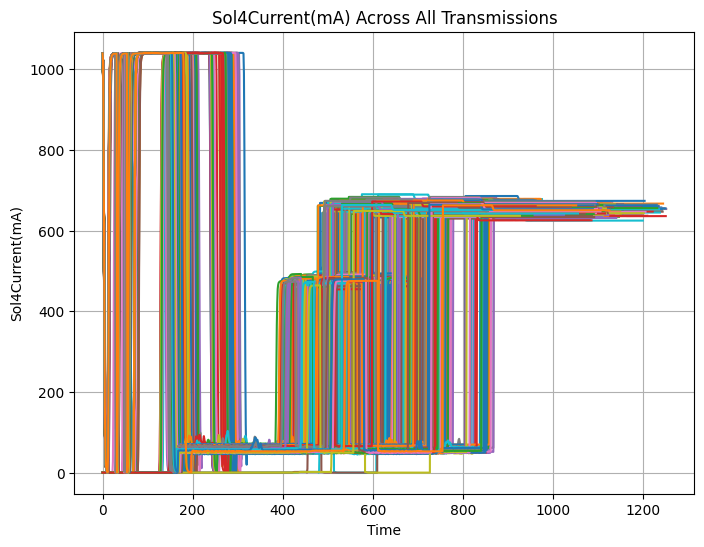} &
\includegraphics[width=0.9\linewidth]{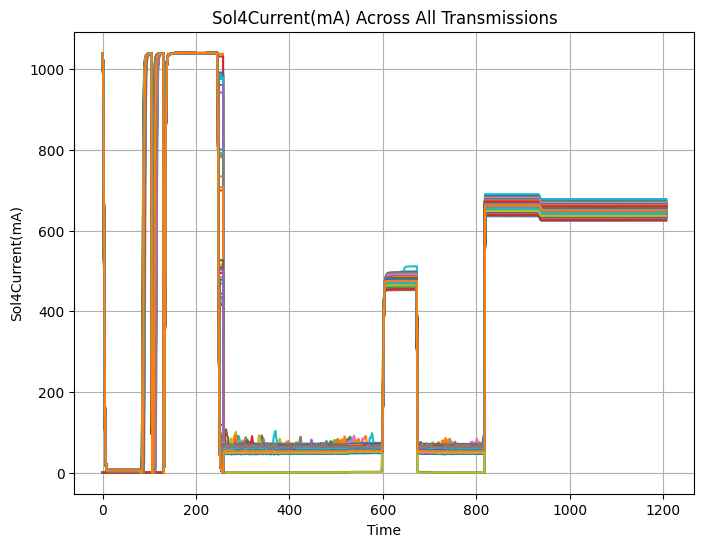} \\
\includegraphics[width=0.9\linewidth]{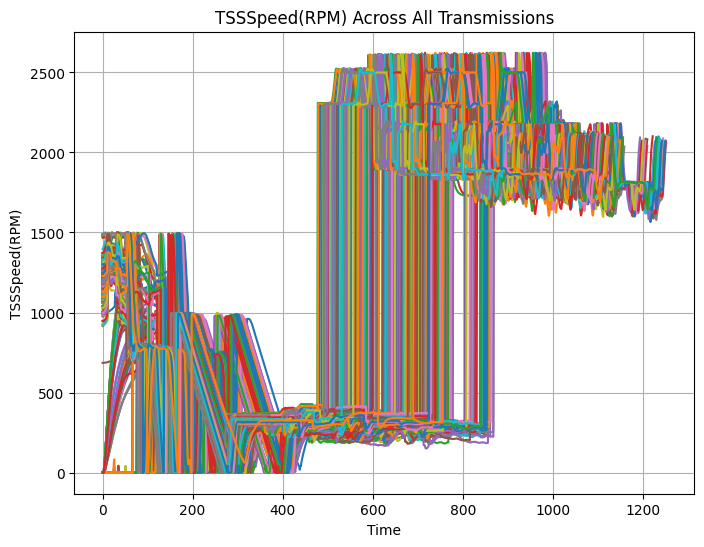} &
\includegraphics[width=0.9\linewidth]{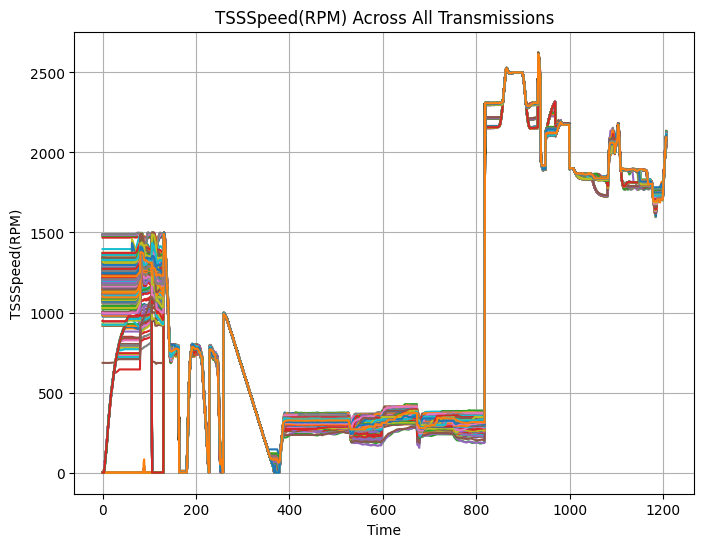} \\[-1em]
\end{tabular}
\caption{Signal traces (Part 2): before vs.\ after DTW.}
\label{fig:dtw-part2}
\end{figure}

\begin{figure}[!htbp]
\centering
\setlength{\tabcolsep}{6pt}
\begin{tabular}{@{}p{.48\linewidth}p{.48\linewidth}@{}}
\textbf{Before applying DTW} & \textbf{After applying DTW} \\[0.0em]
\includegraphics[width=0.9\linewidth]{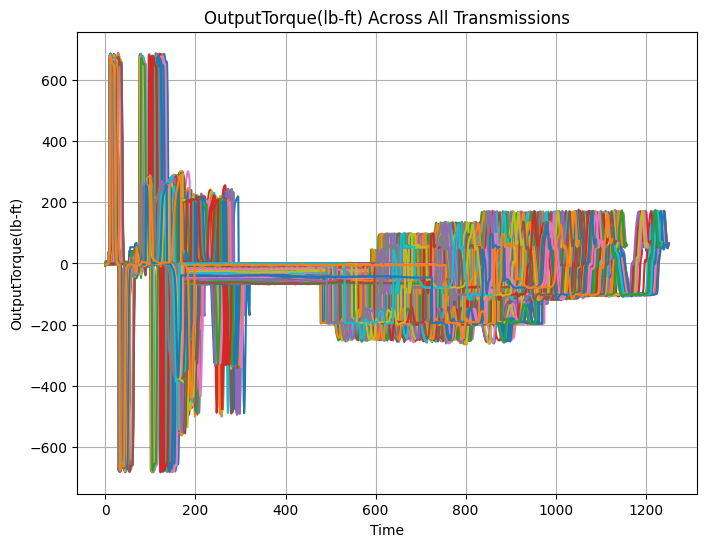} &
\includegraphics[width=0.9\linewidth]{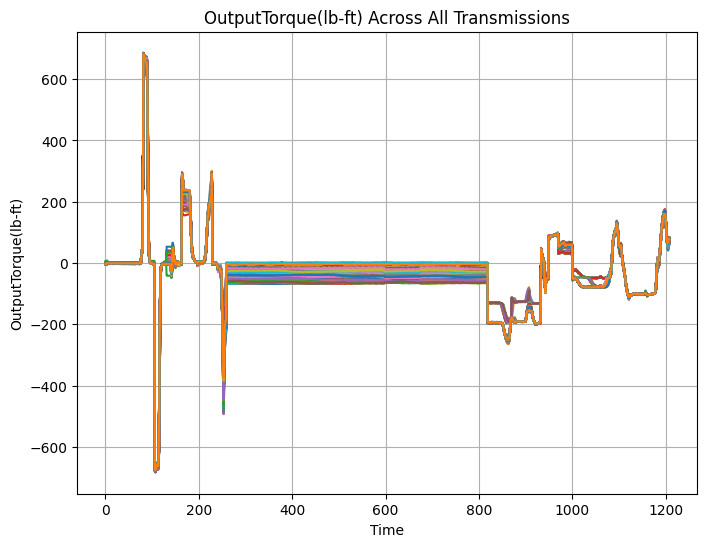} \\[0.0em]
\includegraphics[width=0.9\linewidth]{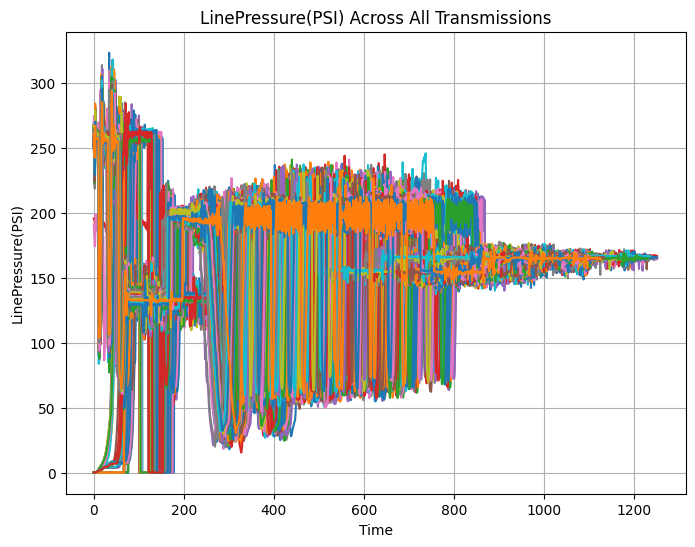} &
\includegraphics[width=0.9\linewidth]{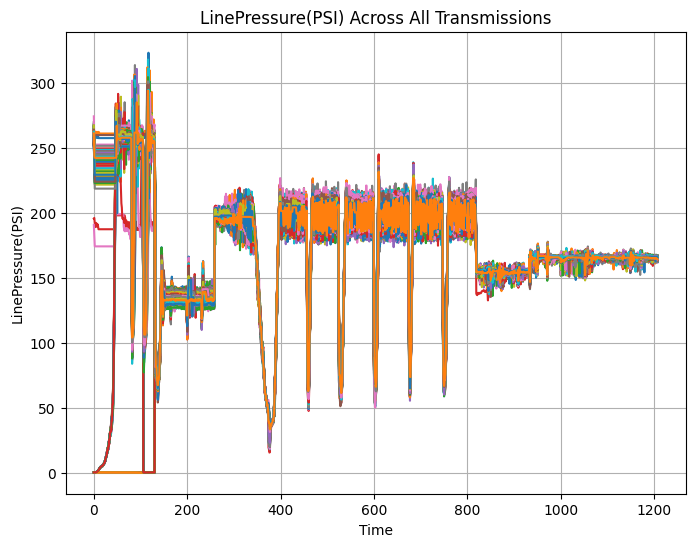} \\[0.0em]
\includegraphics[width=0.9\linewidth]{Images/before_after_dtw/unaligned/tssspeed.png} &
\includegraphics[width=0.9\linewidth]{Images/before_after_dtw/aligned/TSSSPEED.png} \\[0.2em]
\includegraphics[width=0.9\linewidth]{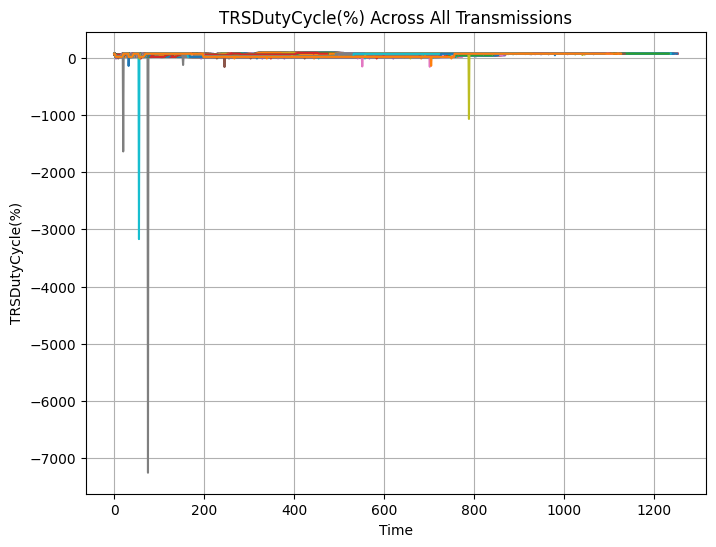} &
\includegraphics[width=0.9\linewidth]{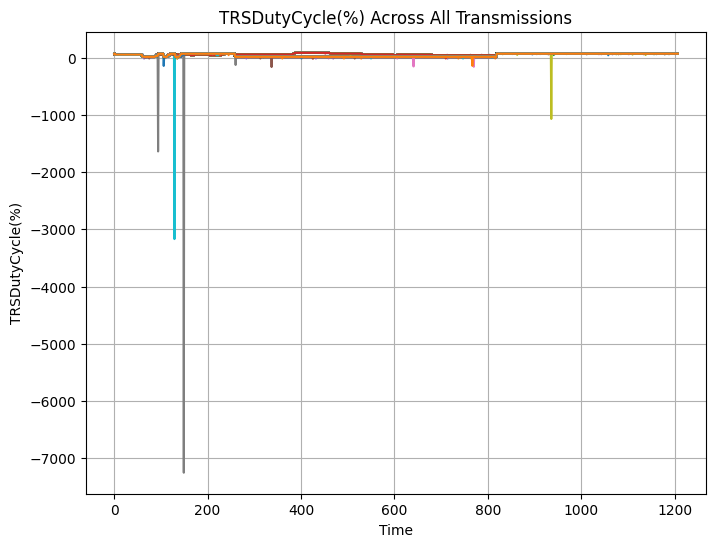} \\[-1em]
\end{tabular}
\caption{Signal traces (Part 3): before vs.\ after DTW.}
\label{fig:dtw-part3}
\end{figure}

For our dataset, we align all of the transmissions to one of the transmissions in the dataset. The choice of the tranmission does not matter but we provide the exact one in the code later for the purpose of exact reproducibility. After aligning all of the transmissions to the reference one, the length of each signal would be 1208. 
In Figs.~\ref{fig:dtw-part1},~\ref{fig:dtw-part2},~\ref{fig:dtw-part3} all signals are shown before and after DTW. Looking at ConvOutPressur signal, it can be seen that there exist two clusters of transmissions. One cluster has an approximate constant value along the time. We reported this to the company and concluded that these need to be diagnosed by the company since their behavior is different from what is expected. Thus we remove these ones, 85 out of 472 and our remaining set will shrink to 387 transmissions.


\subsection{Input Output Relationship - Signal De-correlation}
\begin{table}[!htbp]
\centering
\caption{Sensor input–output relationships.}
\label{tab:input_output}
\begin{tabular}{ll}
\toprule
\textbf{Output Sensor} & \textbf{Input Sensor} \\
\midrule
OutputTorque & InputTorque \\
OutputSpeed     & InputSpeed  \\
OutputSpeed  & TSSSpeed    \\
\bottomrule
\end{tabular}
\end{table}

\begin{figure}[!htbp]
\centering
\includegraphics[width=\textwidth]{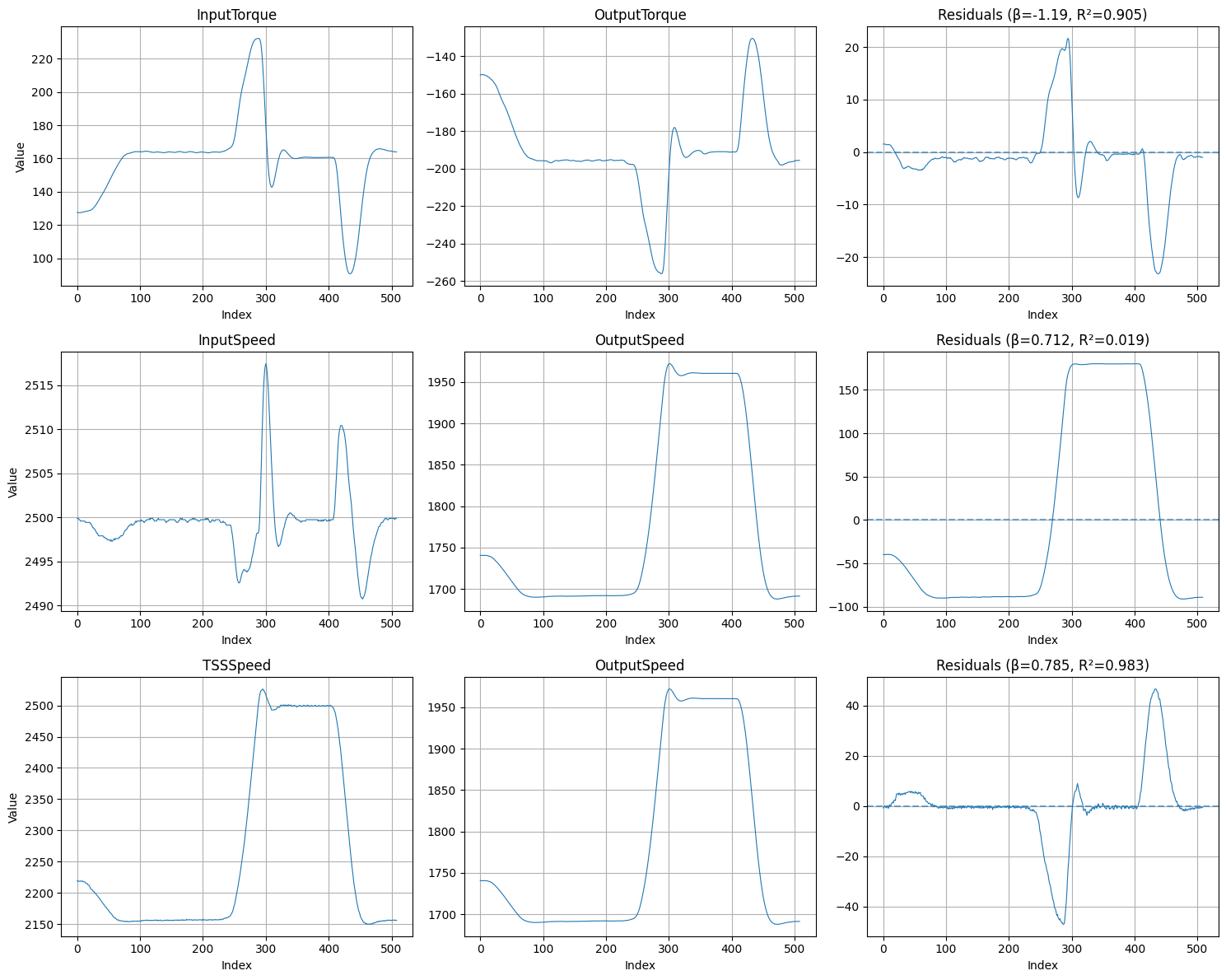}
\caption{Phase 100 of a sample transmission. Columns from left to right shows independent variable, dependent variable and the residual. The name of the sensor signal is in each plot's title.}
\label{Fig:input_output}
\end{figure}

\begin{figure}[!htbp]
\centering

\includegraphics[width=0.32\linewidth]{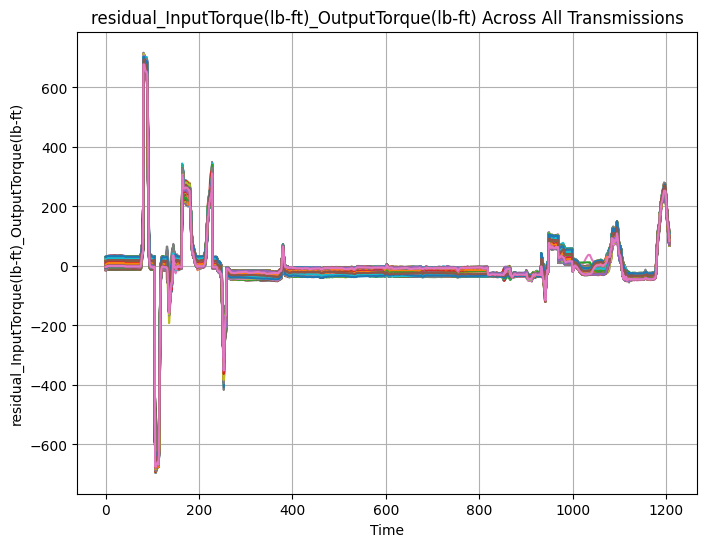}
\includegraphics[width=0.32\linewidth]{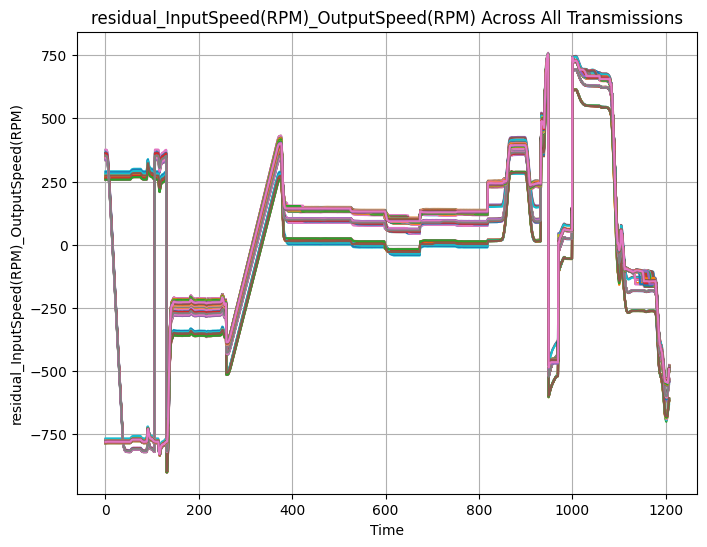}
\includegraphics[width=0.32\linewidth]{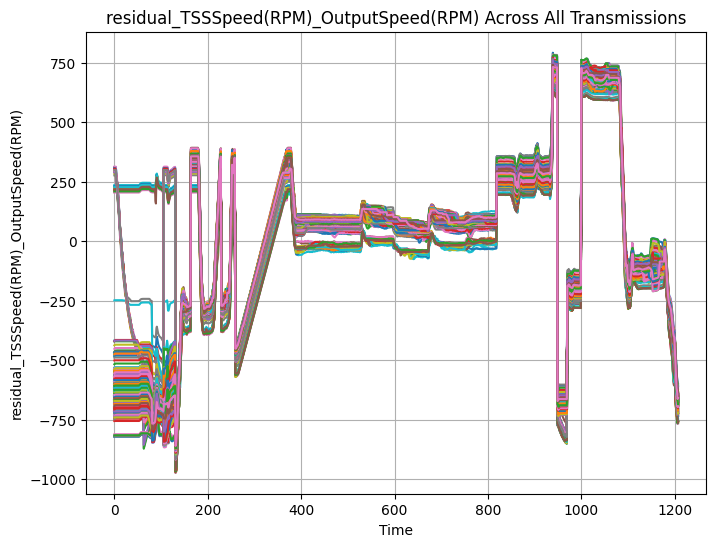}

\caption{Residual signals after regression adjustment. 
Left: Residual of OutputTorque given InputTorque. 
Middle: Residual of OutputSpeed given InputSpeed. 
Right: Residual of OutputSpeed given TSSSpeed.}
\label{fig:residuals}
\end{figure}

Some of the signal pairs as shown in Tab.~\ref{tab:input_output} have a strong correlation. For the monitoring to be more effective, one needs to de correlate the signals. Currently in the company, they use the Euclidean distance between two signals and use that as the new signals used in monitoring. This is problematic because it does not take the complex relationship and different scales of signals into account. Thus we use a regression model and use the residuals as the new signal to monitor. The advantages of this are threefold. First, it includes the information not explained by independent signal. Second, it automatically decorrelates the signal and third, it is not sensitive to the scales of input or output. This scale can come from the external driver increading the input signal, e.g, InputSpeed. In Fig.~\ref{Fig:input_output}, one transmission signal pairs are shown to better visualize the relationship between these signals. 
Fig.~\ref{fig:residuals} shows the residual signals across all transmissions. Since the regression was applied after DTW, the residual signals do not need any further alignment. 



\subsection{Qualitative Results}
After applying the KPCA algorithm to each phase, the resulting feature dimensions for each phase are reported in Table~\ref{tab:feat_dim}.
\begin{table}[ht]
\centering
\caption{Feature dimensions for each phase.}
\label{tab:feat_dim}
\begin{tabular}{l c}
\hline
\textbf{Phase} & \textbf{Feature Dimension} \\
\hline
20.0  & 24 \\
60.0  & 20 \\
70.0  & 42 \\
100.0 & 16 \\
110.0 & 36 \\
\hline
\end{tabular}
\end{table}

\begin{figure}[t]
\centering

\begin{subfigure}[t]{0.48\linewidth}
  \centering
  \includegraphics[width=\linewidth]{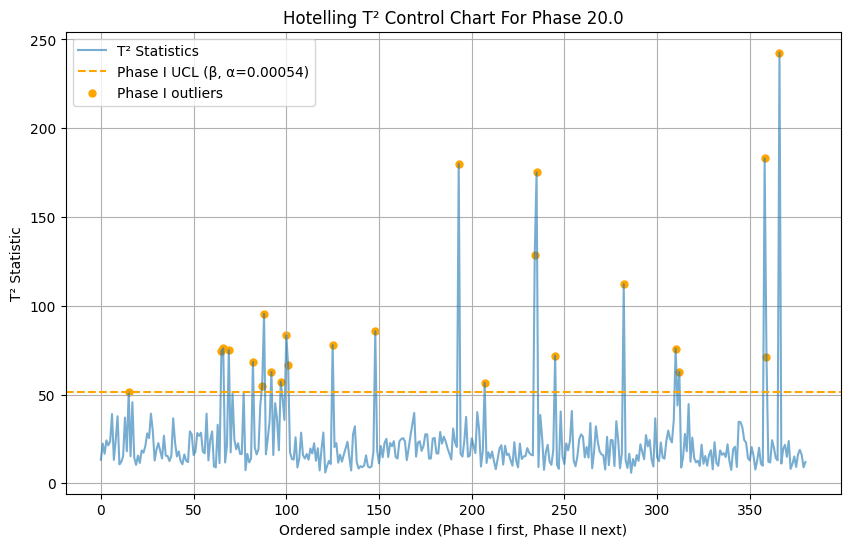}
  \caption*{Phase 20}
\end{subfigure}\hfill
\begin{subfigure}[t]{0.48\linewidth}
  \centering
  \includegraphics[width=\linewidth]{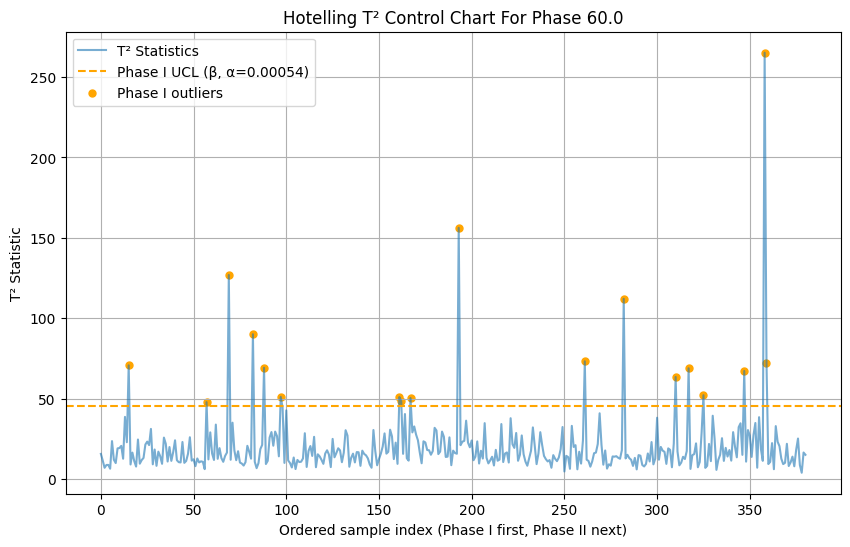}
  \caption*{Phase 60}
\end{subfigure}

\vspace{0.5em}

\begin{subfigure}[t]{0.48\linewidth}
  \centering
  \includegraphics[width=\linewidth]{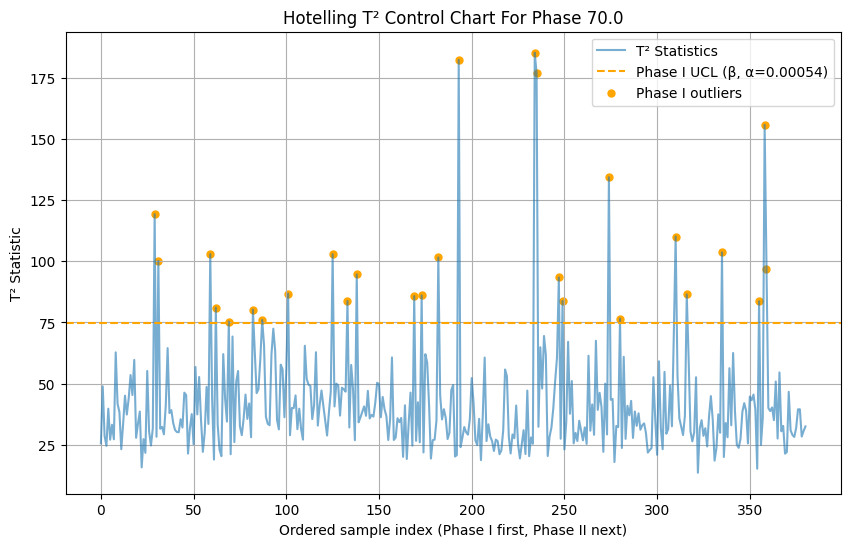}
  \caption*{Phase 70}
\end{subfigure}

\vspace{0.5em}

\begin{subfigure}[t]{0.48\linewidth}
  \centering
  \includegraphics[width=\linewidth]{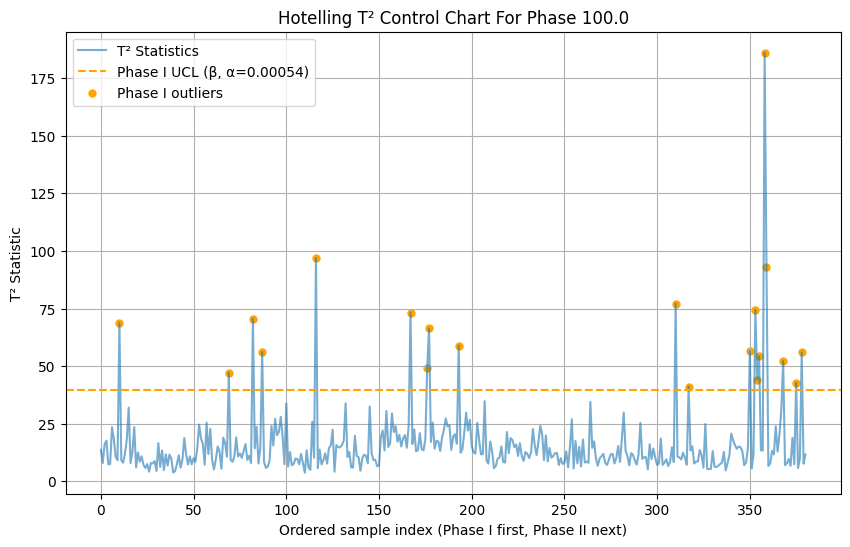}
  \caption*{Phase 100}
\end{subfigure}\hfill
\begin{subfigure}[t]{0.48\linewidth}
  \centering
  \includegraphics[width=\linewidth]{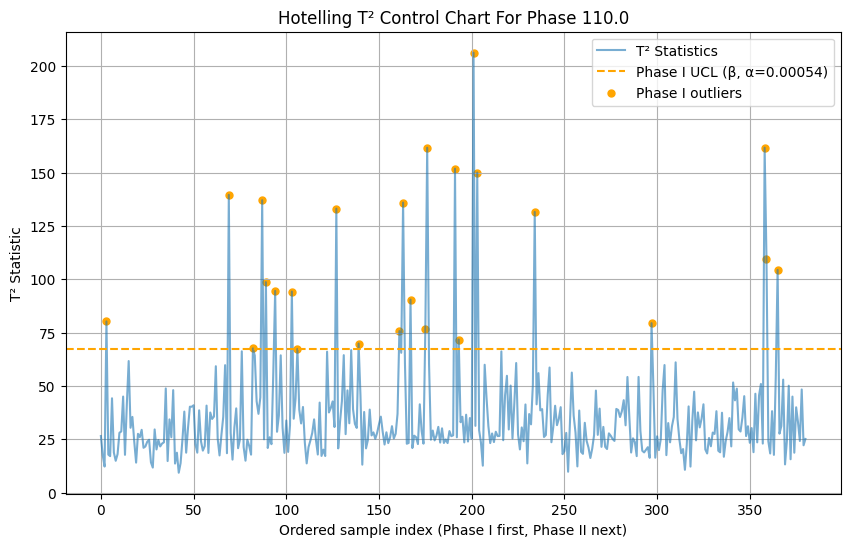}
  \caption*{Phase 110}
\end{subfigure}

\caption{Control chart of the statistics for each phase and control limit. Detected anomalies are highlighted.}
\label{fig:monitoring_plots}
\end{figure}

The Hotelling’s $T^2$ results for each phase are shown in Figure~\ref{fig:monitoring_plots}. To achieve an overall Type I error rate of $0.0027$, corresponding to a three-sigma limit, the Type I error rate for each individual control chart should be set to
$ \alpha = 1-(1-\gamma)^{1/5}
 = 1-(1-0.0027)^{1/5}
 \approx 0.00054.$

\begin{figure}[!htbp]
\centering
\setlength{\tabcolsep}{6pt}
\begin{tabular}{@{}p{.48\linewidth}p{.48\linewidth}@{}}
\includegraphics[width=0.9\linewidth]{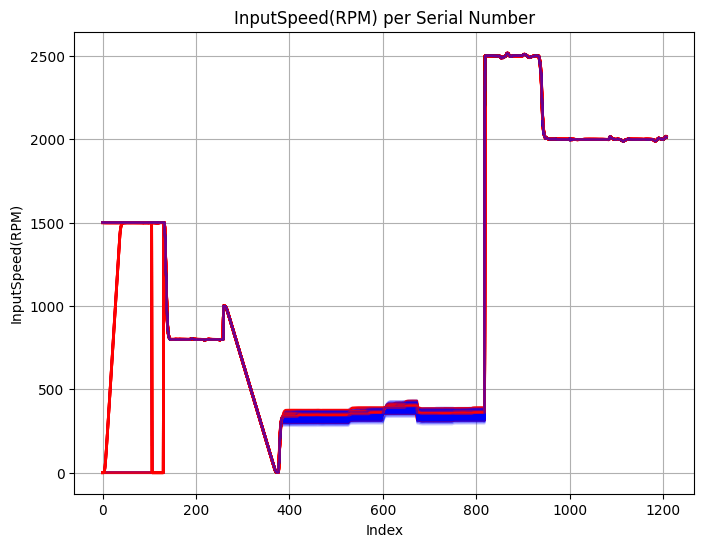} &
\includegraphics[width=0.9\linewidth]{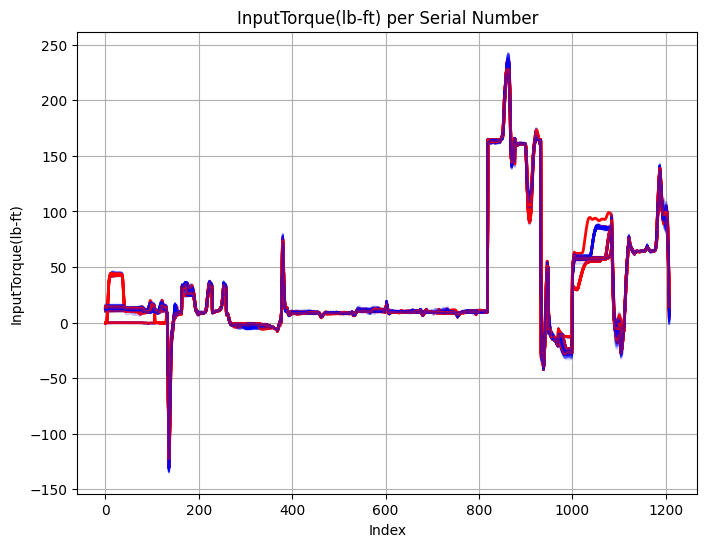} \\[0.6em]
\includegraphics[width=0.9\linewidth]{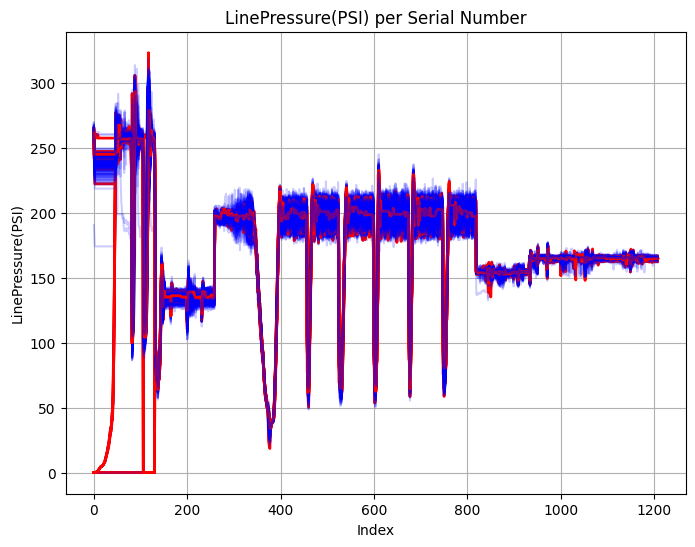} &
\includegraphics[width=0.9\linewidth]{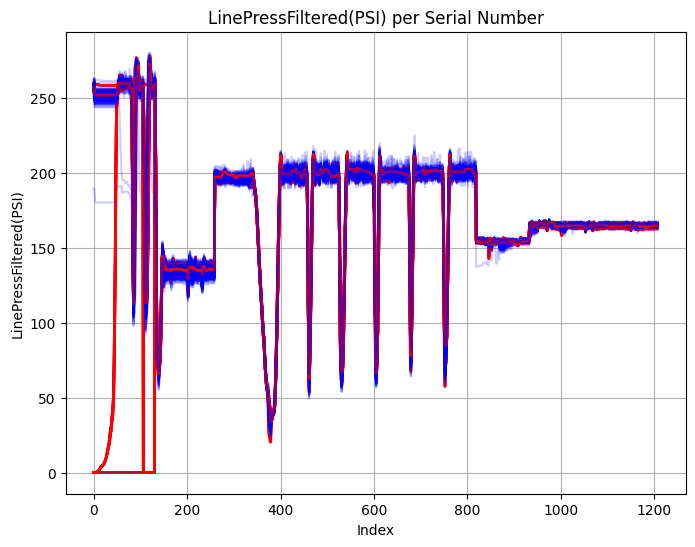} \\[0.6em]
\includegraphics[width=0.9\linewidth]{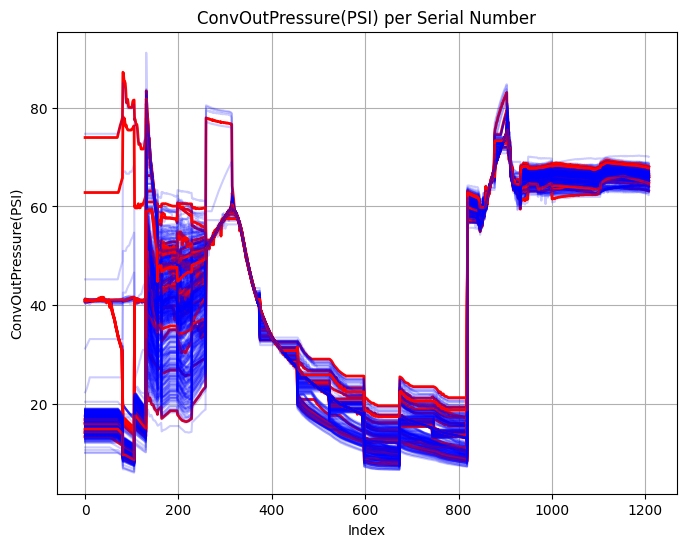} &
\includegraphics[width=0.9\linewidth]{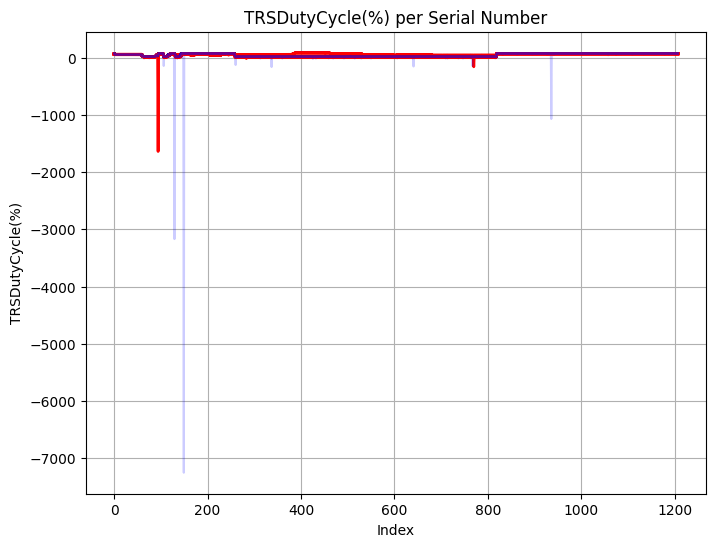} \\
\end{tabular}
\caption{Qualitative results (Part 1): detected anomalies are highlighted in red.}
\label{fig:qual-part1}
\end{figure}

\begin{figure}[!htbp]
\centering
\setlength{\tabcolsep}{6pt}
\begin{tabular}{@{}p{.48\linewidth}p{.48\linewidth}@{}}
\includegraphics[width=0.9\linewidth]{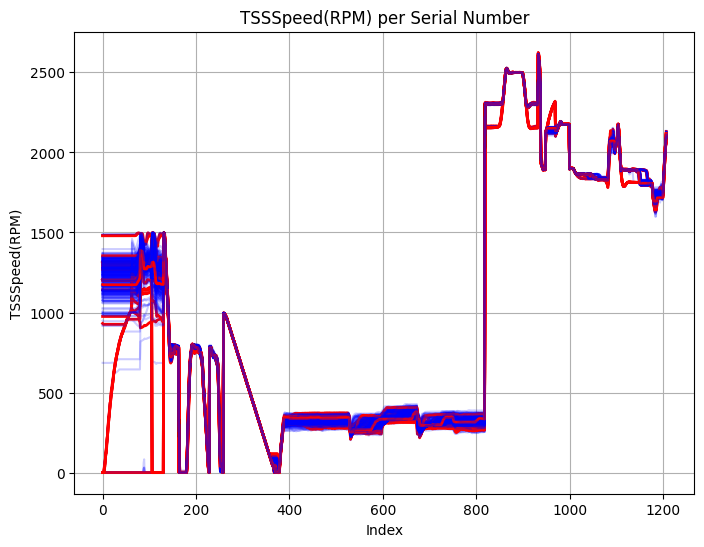} &
\includegraphics[width=0.9\linewidth]{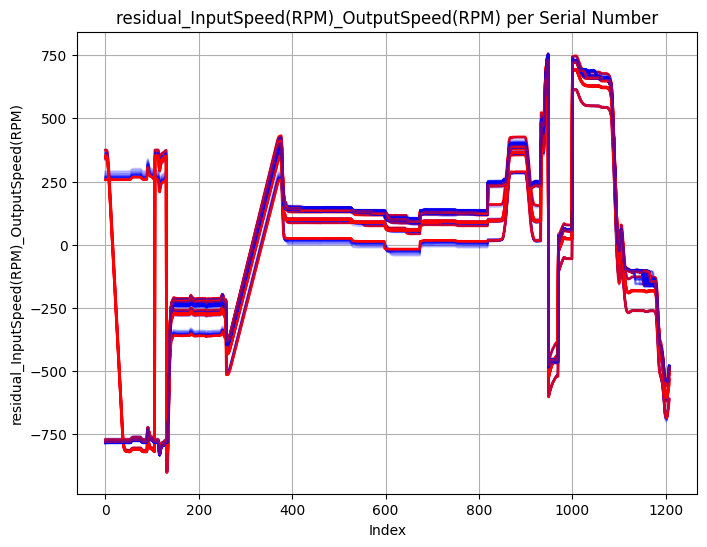} \\[0.6em]
\includegraphics[width=0.9\linewidth]{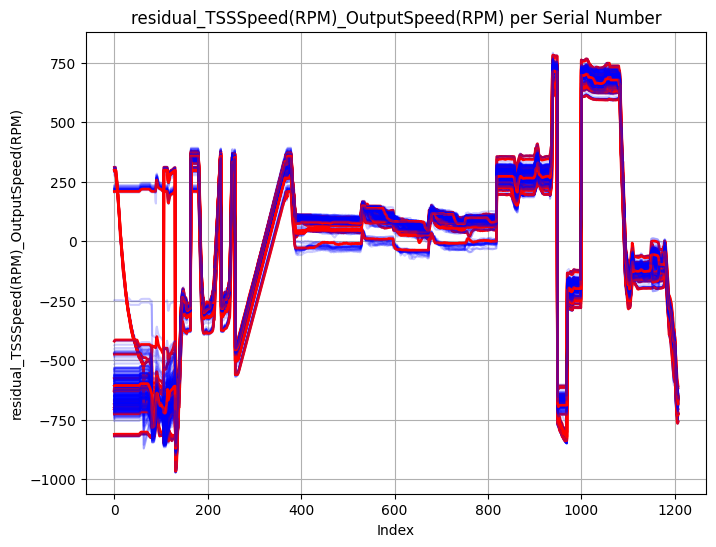} &
\\
\end{tabular}
\caption{Qualitative results (Part 2): detected anomalies are highlighted in red.}
\label{fig:qual-part2}
\end{figure}

To verify the detected anomalies visually, we highlight the top five percent anomalies detected by Alg.~\ref{alg:ranking}. Some signals are demonstrated in Fig.~\ref{fig:qual-part1}, \ref{fig:qual-part2}. The signals are plotted across all of phases combined. As it can be seen, the visually different signals are detected by our algorithm successfully.

\subsection{Quantitative Results on Phase 90 Dataset}

Because our model is applied at the EoL stage after all conventional methods have already been used to detect anomalous transmissions, there are two general ways to evaluate its effectiveness on transmission data. The first is to deploy the transmission in production and then observe whether it is later returned to the company under warranty, which would indicate that it was anomalous. The second is to have an expert review the transmission’s sensor profiles and visually inspect them for abnormal patterns.

Neither of these approaches is effective for our dataset. First, the dataset is new, and insufficient time has passed for enough warranty information to become available. Second, the phases we study are not the standard phase data used by conventional EoL methods for anomaly detection. As a result, in discussions with the company, we found that no established expert evaluation protocol exists for these phases that could provide ground-truth labels. Instead, Phase 90 is a well-documented phase within the company. 

To assess the applicability of our method, we use the Phase 90 transmission data, whose summary is provided in Table~\ref{tab:dataset_summary_phase_90}. The first subset consists of 3,174 transmission samples collected between June 29, 2023 and November 27, 2023, and includes four confirmed anomalous transmissions. The second subset contains 1,545 transmissions of the same type, collected during a different time period that is unknown to us, and does not include any confirmed anomalies. To evaluate the generalization of the proposed method across different transmission types, we also test it on a third dataset consisting of 2,131 samples collected between August 24 and August 30, 2023, which contains three confirmed anomalies.

For this phase, we use steps 120, 160, 121, 161, 122, 182, 124, 164, 125, 165, and 126. We construct a separate control chart for each step, allowing detected anomalies to be traced back to the specific step or steps responsible for the abnormal behavior. For feature extraction, we retain three principal component scores.
\begin{table}[htbp]
\centering
\caption{Summary of the phase 90 transmission datasets used in this study.}
\label{tab:dataset_summary_phase_90}
\renewcommand{\arraystretch}{1.2}

\begin{tabularx}{\linewidth}{
  @{}
  >{\raggedright\arraybackslash}X
  >{\centering\arraybackslash}X
  >{\raggedright\arraybackslash}X
  >{\centering\arraybackslash}X
  @{}
}
\toprule
\textbf{Transmission Type} &
\textbf{Number of Data Points} &
\textbf{Data Collection Period} &
\textbf{Confirmed Anomalies} \\
\midrule
ML3P\_FA & 3174 & 2023-06-29 to 2023-11-27 & 4 \\
ML3P\_FA & 1545 & Not available            & 0 \\
ML3P-PA  & 2131 & 2023-08-24 to 2023-08-30 & 3 \\
\bottomrule
\end{tabularx}
\end{table}

The Phase 90 dataset offers two main advantages. First, it includes confirmed failed transmissions for which warranty claims were activated. Second, expert feedback on the detected anomalies was available. However, because the labeling process was time-consuming, the company experts were only able to provide feedback on the transmissions identified as anomalous by our method. 

\begin{landscape}
\small
\setlength{\tabcolsep}{4pt}
\begin{longtable}{c c c c c c c c c c c}
\caption{Result verification with the current model. The columns display the control chart statistics, where “IC” denotes an in-control observation and numerical entries indicate out-of-control statistics. The company model column corresponds to the predictions made by the state-of-the-art model currently used by the company. The confirmed column identifies transmissions that were previously rejected based on warranty claims, while the worth flagging column represents expert feedback on whether a transmission should be flagged as anomalous.}
\label{tab:result_phase_90_row_1}\\
\toprule
\textbf{Transmission Serial Number}  & \textbf{120} & \textbf{160} & \textbf{161} & \textbf{182} & \textbf{124} & \textbf{164} & \textbf{165} & \textbf{Company} & \textbf{Confirmed} & \textbf{Worth Flagging} \\
\midrule
\endfirsthead

\toprule
\textbf{Transmission Serial Number}  & \textbf{120} & \textbf{160} & \textbf{161} & \textbf{182} & \textbf{124} & \textbf{164} & \textbf{165} & \textbf{Company} & \textbf{Confirmed} & \textbf{Worth Flagging} \\
\midrule
\endhead

Unit-48271 & IC & IC & IC & IC & 2706.86 & IC & IC & No & No & Yes \\
Unit-48371 & IC & IC & IC & IC & IC & 2124.28 & IC & No & No & No \\
Unit-49271 & IC & IC & IC & IC & IC & IC & 2723.93 & No & No & No \\
Unit-48461 & IC & IC & IC & IC & IC & IC & 2810.29 & No & No & Yes \\
Unit-73271 & IC & IC & IC & IC & IC & IC & 2814.91 & No & No & Yes \\
Unit-92271 & IC & IC & IC & 2278.98 & IC & IC & IC & No & No & No \\
Unit-10271 & 2815 & IC & IC & IC & IC & IC & IC & No & No & No \\
Unit-42871 & 2751.83 & IC & IC & IC & IC & IC & IC & Yes & No & No \\
Unit-48871 & IC & 2710.71 & IC & 2815 & IC & IC & IC & No & No & Yes \\
Unit-78271 & IC & IC & IC & IC & IC & IC & IC & No & No & No \\
Unit-49171 & IC & IC & IC & IC & IC & 2678.75 & IC & Yes & No & Yes \\
Unit-48299 & IC & 2688.07 & IC & IC & IC & IC & IC & No & No & Yes \\
Unit-49971 & IC & 2814.55 & IC & 2814.99 & IC & 2029.37 & IC & Yes & No & Yes \\

Unit-18279 & IC & IC & 2781.07 & IC & IC & 955.07 & IC & No & Yes & - \\
Unit-98291 & IC & IC & 2708.04 & IC & IC & 984.82 & IC & No & Yes & - \\
Unit-42291 & IC & IC & 2814.99 & IC & IC & IC & IC & Yes & Yes & - \\

\bottomrule
\end{longtable}
\end{landscape}
\begin{table}[htbp]
\centering
\caption{Performance comparison between the proposed model and the company model for the first subset.}
\label{tab:comparison_metrics_row1}
\renewcommand{\arraystretch}{1.3}

\textbf{Proposed Model}

\begin{tabular}{c|cc}
\toprule
 & Predicted 1 & Predicted 0 \\
\midrule
Actual 1 & 10 & 0 \\
Actual 0 & 6 & 0 \\
\bottomrule
\end{tabular}

\vspace{0.3cm}

Accuracy = $0.625$, \quad
Precision = $0.625$, \quad
Recall = $1.0$, \quad
F1 = $0.769$

\vspace{0.8cm}

\textbf{Company Model}

\begin{tabular}{c|cc}
\toprule
 & Predicted 1 & Predicted 0 \\
\midrule
Actual 1 & 3 & 7 \\
Actual 0 & 1 & 5 \\
\bottomrule
\end{tabular}

\vspace{0.3cm}

Accuracy = $0.50$, \quad
Precision = $0.75$, \quad
Recall = $0.30$, \quad
F1 = $0.429$

\end{table}
\begin{table}[htbp]
\centering
\caption{Control chart statistics for the second phase 90 subset.}
\label{tab:result_phase_90_row_2}
\renewcommand{\arraystretch}{1.2}
\resizebox{\textwidth}{!}{%
\begin{tabular}{lcccccccccccc}
\toprule
\textbf{Serial} & \textbf{120} & \textbf{160} & \textbf{121} & \textbf{161} & \textbf{122} & \textbf{182} & \textbf{124} & \textbf{164} & \textbf{125} & \textbf{165} & \textbf{126} & \textbf{Worth Flagging} \\
\midrule
Unit-26589 & IC & 1308.25 & IC & IC & IC & IC & IC & 1440.61 & 1254.17 & IC & IC & Yes \\
Unit-54379 & IC & IC & IC & IC & IC & IC & 1369.92 & IC & IC& IC & IC & No \\
Unit-33498 & IC & IC & IC & 1542.86 & IC & IC & 1407.86 & 1542.34 & IC & IC & 1540.70 & Yes \\
Unit-12967 & IC & IC & 1540.85 & IC & IC & IC & IC & IC & IC & 1542.99 & IC & Yes \\
Unit-72491 & 1537.60 & IC & IC & IC &IC & IC & IC & IC & IC & IC& IC& No \\
\bottomrule
\end{tabular}%
}
\end{table}

\begin{table}[htbp]
\centering
\caption{Confusion matrix and performance metrics of our model for the second phase 90 dataset subset.}
\label{tab:confusion_metrics_row_2}
\renewcommand{\arraystretch}{1.3}

\textbf{Model}

\begin{tabular}{c|cc}
\toprule
 & Predicted 1 & Predicted 0 \\
\midrule
Actual 1 & 3 & 0 \\
Actual 0 & 2 & 0 \\
\bottomrule
\end{tabular}

\vspace{0.3cm}

Accuracy = $0.60$, \quad
Precision = $0.60$, \quad
Recall = $1.00$, \quad
F1 = $0.75$
\end{table}

\begin{table}[htbp]
\centering
\caption{Control chart statistics for the third subset of phase 90 dataset. The confirmed column indicates transmissions with warranty claims or historical reject code. For this subset, no company expert feedback was available.}
\label{tab:result_phase_90_row_3}
\renewcommand{\arraystretch}{1.2}
\setlength{\tabcolsep}{4pt}
\small
\resizebox{\textwidth}{!}{%
\begin{tabular}{lccccccc}
\toprule
\textbf{Serial Number} & \textbf{120} & \textbf{160} & \textbf{161} & \textbf{164} & \textbf{165} & \textbf{182} & \textbf{Confirmed} \\
\midrule
Unit-58341 & 1797.995 & IC & IC & IC & IC & IC & No \\
Unit-59183 & IC & 1278.996 & IC & IC & IC & IC & No \\
Unit-63517 & IC & IC & IC & 1266.76 & IC & IC & No \\
Unit-51829 & IC & IC & 1429.117 & IC & IC & IC & No \\
Unit-10582 & IC & IC & IC & IC & 807.0313 & IC & No \\
Unit-54986 & IC & IC & IC & IC & 1787.933 & IC & No \\
Unit-16597 (reject\_1)        & IC & IC & IC & IC & IC & 1391.137 & Yes \\
Unit-54986 (reject\_2)        & 1792.414 & 1797.977 & 1798 & 1798.001 & IC & 1739.196 & Yes \\
Unit-84621 & IC & IC & IC & IC & IC & 6466.292 & No \\
Unit-49782 & 441.4047 & IC & IC & IC & 917.6367 & IC & No \\
Unit-98053 & IC & IC & IC & 2707.024 & IC & IC & No \\
Unit-10582 (reject\_0)        & IC & 555563.2 & 1.21E+07 & IC & IC & IC & Yes \\
\bottomrule
\end{tabular}%
}
\end{table}

For the first subset, we set the threshold so that 0.005041 of the samples are identified as anomalous. The results are reported in Table~\ref{tab:result_phase_90_row_1}. For each transmission, the control chart statistic is computed for every step. If the statistic falls below the upper control limit, it is labeled as IC; otherwise, its numerical value is reported. The company model column indicates whether the corresponding transmission is detected by the current state-of-the-art model used by the company. The confirmed column indicates whether the transmission has an associated warranty claim or a historical anomaly code. The worth flagging column reflects expert assessment of whether the transmission should be considered anomalous after manual inspection.

If we treat all transmissions flagged by our method as the evaluation set and define the ground-truth label using the last two columns, we can compare the confusion matrix, as well as the accuracy, precision, recall, and F1 score, of our method against the current company model. These results are shown in Table~\ref{tab:comparison_metrics_row1}, which demonstrates that our model substantially outperforms the benchmark in terms of accuracy, recall, and F1 score. In particular, our model achieves an F1 score of 0.769, compared with 0.429 for the benchmark.


For the second subset, 0.3236\% of the entire dataset is identified as anomalous. The results are presented in Table~\ref{tab:result_phase_90_row_2}, which makes it possible to determine which steps exhibit abnormal behavior. This information can support future root-cause analysis by the company. Similar to the first subset, performance metrics are reported in Table~\ref{tab:confusion_metrics_row_2}, where our model achieves an accuracy of 60\% and an F1 score of 0.75. In these tables, serial numbers were de-identified and replaced with anonymized research IDs to preserve confidentiality while maintaining sample traceability throughout the analysis.

For the third subset, the results are shown in Table~\ref{tab:result_phase_90_row_3}. For this dataset, we were unable to obtain company feedback; however, our method successfully detects all three confirmed anomalies while flagging 0.0056 of the entire dataset as anomalous.

Practitioners can adjust the proportion of detected anomalies in the dataset through the choice of Type I error or related design parameters. In general, increasing this proportion tends to reduce false negatives at the expense of more false positives. The appropriate balance depends on business rules, operational needs, and the company’s tolerance for false positives and false negatives.

\subsection{Online Weight Update for Alg.\ref{alg:ranking}}
\begin{figure}[!htbp]
    \centering
    \includegraphics[width=\linewidth]{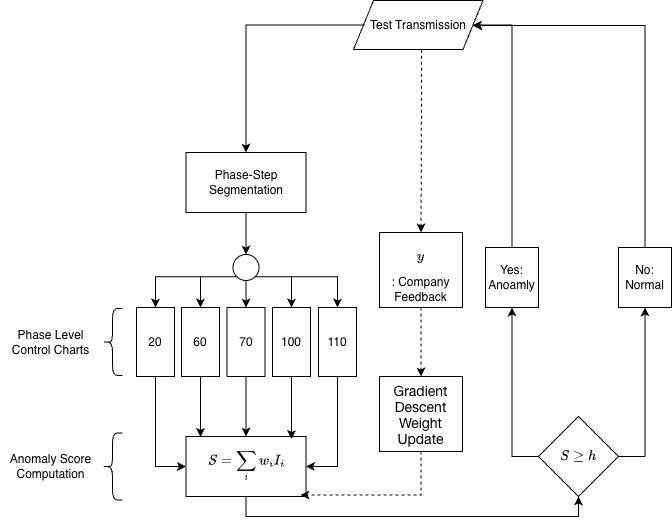}
    \caption{An overview of the data-Driven parameters update in our decision process.}
    \label{fig:feedback}
\end{figure}

As we discussed for the case of unsupervised anomaly detection when no ground truth label is available, one can use the developed methodologies and choose the top anomalies based on historical data of warranty rate. Due to the confidentiality, we cannot disclose this number for Ford Motor Company. However, the algorithm can be deployed in a production environment and feedback will be received regarding the transmission. For example, a transmission might be considered normal if after one year no warranty claim was filed for it. Thus in a streaming manner, ground truth labels are going to be received. 

Over time, our aim is to learn the optimal phase weights from Alg. \ref{alg:ranking}. To do that we reconstruct an anomaly detection model as a binary classifier and then update the weights based on gradient descent optimization algorithm. Current anomaly score computation shows relative rank information for detected anomalies to find the top anomalies. This score, however, does not have a range thus we feed it to a sigmoid function for the output to be between 0 and 1. Thus the output will be an anomaly probability. 

Additionally, we update the weights based on the feedback for all of the serial numbers as some that were not detected as anomaly in any of the phases might be anomalous or vice versa. Thus the input to the algorithm will be all deployed transmissions. 

To learn the optimal weights, we use a binary cross entropy loss and use gradient descent algorithm to update the weights. For weight updates, we use a fixed learning rate. More sophisticated gradient based algorithms such as Adam \cite{kingma2014adam}, mini batch gradient descent, stochastic mini batch gradient descent can be used.

As shown in Alg.~\ref{alg:online-weight-learning-warranty}, the classifier is a binary classifier with sigmoid function enforcing the output to be between 0 and 1. For the loss function, binary cross entropy is used to enforce the predicted probabilities follow the ground truth labels. The gradient of the loss function with respect to each weight is also calculated and parameter updates using a fixed learning rate is shown. Note that after each update with any ground truth, all weights are divided by their total summation so their value would add up to one again. 
\begin{algorithm}[H]
\caption{Online Weight Learning from Warranty Feedback}
\begin{algorithmic}[1]
\Require Phases indexed by $i=1,\dots,m$; learning rate $\eta>0$, phase specific weights $w_i$, anomaly indicator for each phase $I_i$, phase-level threshold $T_i$, phase-level anomaly score $S_i$
\Statex

\State \textbf{Classifier (binary):}
\State Define the logit
\[
z \;=\; \sum_{i=1}^{m} w_i\, I_i\, (S_i - T_i),
\]
and the predicted probability
\[
\hat{y} \;=\; \sigma(z) \;=\; \frac{1}{1+e^{-z}}, \qquad \hat{y}\in(0,1).
\]
\Statex

\State \textbf{Loss (binary cross-entropy):}
For a warranty feedback label $y\in\{0,1\}$, define
\[
\mathcal{L}(y,\hat{y}) \;=\; -\Big( y\log(\hat{y}) + (1-y)\log(1-\hat{y}) \Big).
\]
\Statex
\State \textbf{Gradients:}
Using $\frac{\partial \mathcal{L}}{\partial z} = \hat{y}-y$ and $\frac{\partial z}{\partial w_k}= I_k(S_k-T_k)$,
\[
\frac{\partial \mathcal{L}}{\partial w_k}
\;=\;
(\hat{y}-y)\, I_k\, (S_k-T_k),
\qquad k=1,\dots,m.
\]
\State \textbf{Online update for a transmission with label $y$:} \For{$k=1$ to $m$}
    \State $w_k \leftarrow w_k - \eta\,(\hat{y}-y)\, I_k\, (S_k-T_k)$
\EndFor
\State \textbf{Weight normalization (keep $\sum_i w_i$ constant, e.g., 1):}
$W \leftarrow \sum_{i=1}^{m} w_i$
\For{$i=1$ to $m$}
    \State $w_i \leftarrow \dfrac{w_i}{W}$
\EndFor
\end{algorithmic}
\label{alg:online-weight-learning-warranty}
\end{algorithm}

\section{Conclusion}
This study presents a comprehensive data-driven framework for monitoring transmission systems using high-dimensional EoL testing data. The proposed approach integrates a structured preprocessing pipeline, including data cleaning, cycle removal, smoothing, and temporal alignment, to address inconsistencies commonly observed in industrial testing environments. To account for physical dependencies among signals, an input--output modeling step based on regression is introduced, enabling the extraction of residual signals that better reflect abnormal behavior. These processed signals are then transformed into a compact representation using Kernel Principal Component Analysis (KPCA), capturing the underlying nonlinear structure of the system. Finally, a multivariate statistical monitoring scheme based on Hotelling’s $T^2$ statistic is employed for anomaly detection, complemented by a weighted scoring mechanism to prioritize potentially faulty transmissions.

The effectiveness of the proposed framework is demonstrated through a real-world case study using production data from an automotive manufacturing setting. The results show that the methodology can reliably identify anomalous transmissions while accounting for temporal variability, signal coupling, and high dimensionality. By combining nonlinear feature extraction with statistical process monitoring, the framework provides a practical and interpretable solution for quality control in complex manufacturing systems.

For the future research, one can extend the work to a semi-supervised setting where some reference anomaly data is available and the model is tuned on them to decrease its false positive rate. Additionally, the current implementation of the model consists of two stages. In the first one, low dimensional features are extracted and in the second stage monitoring is done. One can also extend the model to be end-to-end in one step.

\section*{Gen AI Use}
ChatGPT was used for grammar checking, text polishing, and code development. All ideas presented in this work were developed by the authors. Generative AI tools were used only to assist with code implementation and to polish or improve the written text of the paper. All outputs produced by generative AI were thoroughly reviewed by the authors. 

\section*{Acknowledgment}
\bibliographystyle{chicago}
\bibliography{IISE_Trans}
\end{document}